\documentclass[dvipsnames,format=sigconf,anonymous=false,review=false]{acmart} 

\def\vec#1{\mathchoice{\mbox{\boldmath$\displaystyle#1$}}
{\mbox{\boldmath$\textstyle#1$}}
{\mbox{\boldmath$\scriptstyle#1$}}
{\mbox{\boldmath$\scriptscriptstyle#1$}}}
\usepackage{booktabs} % For formal tables
\usepackage{amsmath}
\usepackage{algorithm}
\usepackage[noend]{algpseudocode}
\usepackage[T1]{fontenc}
\usepackage{tabularx}
\usepackage{multirow}
\usepackage{array}
\usepackage{pbox}

\usepackage{color,soul}
\usepackage{nameref}

\usepackage{pgfplots}
\usepackage{graphicx}
\usepackage{subcaption}
\usepackage{tikz}
\usetikzlibrary{intersections, pgfplots.fillbetween}
\usepackage{siunitx}

\usepackage{url}
\usepackage{bm}

\definecolor{darkgreen}{rgb}{0.30, 0.50, 0.0}

\makeatletter
\renewcommand{\ALG@name}{Pseudocode}
\makeatother

\newcolumntype{L}[1]{>{\raggedright\let\newline\\\arraybackslash\hspace{0pt}}m{#1}}
\newcolumntype{C}[1]{>{\centering\let\newline\\\arraybackslash\hspace{0pt}}m{#1}}
\newcolumntype{R}[1]{>{\raggedleft\let\newline\\\arraybackslash\hspace{0pt}}m{#1}}

\graphicspath{{images/}}

\copyrightyear{2025}
\acmYear{2025}
\setcopyright{cc}
\setcctype{by}
\acmConference[GECCO '25]{Genetic and Evolutionary Computation
Conference}{July 14--18, 2025}{Malaga, Spain}
\acmBooktitle{Genetic and Evolutionary Computation Conference (GECCO '25),
July 14--18, 2025, Malaga, Spain}\acmDOI{10.1145/3712256.3726410}
\acmISBN{979-8-4007-1465-8/2025/07}

\hypersetup{draft}
\begin{document}
	\title[On Revealing the Hidden Problem Structure Using Walsh Coefficient Influence]{On Revealing the Hidden Problem Structure in Real-World and Theoretical Problems Using Walsh Coefficient Influence}

 	\author{Michal W. Przewozniczek}
 	\affiliation{
            %\institution{Dep. of Systems and Comp. Networks}
 		\institution{Wroclaw Univ. of Science and Techn.}
 		\city{Wroclaw}
 		\country{Poland} 
 	}
 	\email{michal.przewozniczek@pwr.edu.pl}

        \author{Francisco Chicano}
        \affiliation{
        %\institution{Dep. of Lang. and Comp. Sciences}
          \institution{ITIS Software, University of M\'alaga}
         \city{M\'alaga} 
         \country{Spain} 
         \postcode{29071}
        }
        \email{chicano@uma.es}

        \author{Renato Tin\'os}
        \affiliation{
        %\institution{Dep. of Computing and Mathematics}
        \institution{University of S\~ao Paulo}
        \city{Ribeir\~ao Preto} 
        %\STATE{S\~ao Paulo}
        \country{Brazil}
        \postcode{14040901}
        }
        \email{rtinos@ffclrp.usp.br}

        \author{Jakub Nalepa}
 	\affiliation{
 	%\institution{Dep. of Algorithmics and Software}
 	\institution{Silesian Univ. of Technology/KP Labs}
 	\city{Gliwice}
 	\country{Poland} 
 	}
 	\email{jakub.nalepa@polsl.pl}

        \author{Bogdan Ruszczak}
        \affiliation{
        %\institution{Department of Informatics}
        \institution{Opole Univ. of Technology/KP Labs}
        \city{Opole}
        \country{Poland} 
        }
 	\email{b.ruszczak@po.edu.pl}

\author{Agata M. Wijata}
        \affiliation{
        %\institution{Dep. of Medical Informatics and AI}
        \institution{Silesian Univ. of Technology/KP Labs}
        \city{Gliwice}
        \country{Poland} 
        }
        \email{agata.wijata@polsl.pl}        

        %\author{Marcin M. Komarnicki}
 	%\affiliation{
 	%\institution{Dep. of Systems and Comp. Networks}
 	%	\institution{Wroclaw Univ. of Science and Techn.}
 	%	\city{Wroclaw}
 	%	\country{Poland} 
 	%}
 	%\email{marcin.komarnicki@pwr.edu.pl}

	\renewcommand{\shortauthors}{Michal W. Przewozniczek et al.}

	\begin{abstract}
        Gray-box optimization employs Walsh decomposition to obtain non-linear variable dependencies and utilize them to propose masks of variables that have a joint non-linear influence on fitness value. These masks significantly improve the effectiveness of variation operators. In some problems, all variables are non-linearly dependent, making the aforementioned masks useless. We analyze the features of the real-world instances of such problems and show that many of their dependencies may have noise-like origins. Such noise-caused dependencies are irrelevant to the optimization process and can be ignored. To identify them, we propose extending the use of Walsh decomposition by measuring variable dependency strength that allows the construction of the weighted dynamic Variable Interaction Graph (wdVIG). wdVIGs adjust the dependency strength to mixed individuals. They allow the filtering of irrelevant dependencies and re-enable using dependency-based masks by variation operators. We verify the wdVIG potential on a large benchmark suite. For problems with noise, the wdVIG masks can improve the optimizer's effectiveness. If all dependencies are relevant for the optimization, i.e., the problem is not noised, the influence of wdVIG masks is similar to that of state-of-the-art structures of this kind.
        \end{abstract}
	
    

	

	
	\maketitle
	
	%\textcolor{red}{\textbf{The supplementary material for this document is available at}}\\
%\textcolor{red}{\url{http://www.przewozniczek.eu/download/weightWalshSUPPL.pdf}}

\section{Introduction}
\label{sec:intro}
%Gray-box optimization and model-building optimizers in black-box optimization use the knowledge about variable dependencies to improve their performance. Two variables may be dependent if their joint influence on the optimized function is non-linear. Thus, such variables shall be processed together by an optimizer. Frequently, such knowledge is employed to propose variation masks \cite{whitleyNext}, e.g., for individual mixing \cite{pxForBinary} or as perturbation masks~\cite{ilsDLED}.\par
Gray-box optimization and model-building optimizers in black-box optimization use the knowledge about variable dependencies to improve their performance. Two variables may be considered dependent if their joint influence on the optimized function is non-linear. Such variables shall be processed together by an optimizer. Therefore, dependencies are employed to construct variation masks \cite{whitleyNext}, e.g., for individual mixing \cite{pxForBinary} or perturbation~\cite{ilsDLED}.\par

To create a high-quality mask, we consider the \textit{problem structure}, i.e., the graph of variable dependencies, to join dependent variables into one cluster. Such mechanisms are effective when dealing with a \textit{strong problem structure} (when the percentage of dependent variable pairs is low). However, if the structure is \textit{weak} (all or almost all variable pairs are dependent), we mainly obtain full masks joining all or almost all variables, which may be useless. To overcome this difficulty, we can estimate the dependency strength to use only the strongest dependencies \cite{GAwLL}. However, it may be hard to decide what threshold shall we use for the mask creation.\par

%Walsh decomposition \cite{heckendorn2002,GrayBoxWhitley,whitleyNext} is frequently used in gray-box optimization to discover variable dependencies before the optimization process. To the best of our knowledge, it has only been employed to differentiate dependent variable pairs from non-dependent ones. Therefore, we aim to extend the usability of Walsh decomposition to obtain the strength of dependencies. To this end, we propose, analyze, and verify experimentally different propositions of Walsh-based dependency strength estimation. On this base, we propose mechanisms that adjust the strength of the mixed individuals. Finally, we integrate all these mechanisms in an optimizer that is shown more effective than the competing optimizers.\par

Walsh decomposition \cite{heckendorn2002,GrayBoxWhitley,whitleyNext} is frequently used in gray-box optimization to discover variable dependencies before the optimization process. To the best of our knowledge, it has only been employed to differentiate dependent from non-dependent variable pairs. Therefore, we extend the usability of Walsh decomposition by obtaining the dependency strength. To this end, we propose, analyze, and experimentally verify different propositions of Walsh-based dependency strength estimation, including the mechanisms that adjust the dependency strength to mixed individuals. Finally, we integrate all these mechanisms in an optimizer that outperforms other state-of-the-art propositions.\par

To improve the understanding of the applicability of our propositions, we thoroughly investigate the structure-related features of toy-sized instances of several real-world problems arising in the field of Machine Learning. We discover that many dependencies may have noise-like origins and have negligible influence on the optimized function value. As such, these existing dependencies can be ignored while constructing variation masks. These observations show that proposing accurate dependency strength measurement may be a game-changer in the optimization of many real-world problems to which gray-box mechanisms can not be applied today. Moreover, in optimization, it is frequent to consider non-linear dependencies. 
Our results show that other dependency types can be significantly more convenient in filtering irrelevant dependencies and obtaining high-quality variation masks.\par

\section{Background}
\label{sec:relWork}

\subsection{Gray-box optimization and Walsh decomposition}
\label{sec:relWork:grayWalsh}
%\textcolor{red}{\textbf{FILL IT UP}}\\
%\textcolor{red}{1. Additive form, say something about overlaps and that the overlapping problems are harder}\\
%\textcolor{red}{2. Walsh form}\\
%\textcolor{red}{\textbf{3. Individual = a summ of Walsh coefficients. Thus, we wish to exchange the fitness-improving values of Walsh coefficients} This issue is super-important for section \ref{sec:PXwdVIG:intutions}}\\
%\textcolor{red}{4. Walsh-based VIG}\\
%\textcolor{red}{4. Partial evaluations, but very shortly because we do not use them in the paper}\\
Gray-box optimization utilizes the knowledge about variable dependencies to leverage the optimization process \cite{whitleyNext}. It frequently focuses on the \textit{k}-bounded problems, which can be represented as the sum of subfunctions taking no more than \textit{k} arguments \cite{transTokBounded}. The \textit{additive form} is convenient for representing such problems: $f(\vec{x})=\sum_{s=1}^{S} f_s(\vec{x}_{I_s})$, where $\vec{x}=(x_1,x_2,\ldots,x_{n})$ is a binary vector of size $n$, $I_s$ are subsets of $\{1,...,n\}$ ($I_s$ do not have to be disjoint), and $S$ is the number of these subsets.\par
%\begin{equation}
 %       \label{eq:additive}
 %       f(\vec{x})=\sum_{s=1}^{S} f_s(\vec{x}_{I_s}),
 %   \end{equation}
 %   where $\vec{x}=(x_1,x_2,\ldots,x_{n})$ is a binary vector of size $n$, $I_s$ are subsets of $\{1,...,n\}$ ($I_s$ do not have to be disjoint), and $S$ is the number of these subsets.\par

Walsh decomposition \cite{heckendorn2002} is useful in decomposing \textit{k}-bounded problems and represent them in as: $f(\vec{x}) = \sum_{i=0}^{2^n-1} w_i \varphi_i(\vec{x})$, where $w_i \in \mathbb{R}$ is the $i$th Walsh coefficient, $\varphi_i(\vec{x}) =(-1)^{\mathbf{i}^\mathrm{T}\vec{x}}$ defines a sign, and $\mathbf{i} \in \{0,1\}^n$ is the binary representation of index $i$.  \par

    %\begin{equation}
    %\small
    %\label{eq:walsh-decomposition}
    %f(\vec{x}) = \sum_{i=0}^{2^n-1} w_i \varphi_i(\vec{x}), 
    %\end{equation}
    %where $w_i \in \mathbb{R}$ is the $i$th Walsh coefficient, $\varphi_i(\mathbf{x}) =(-1)^{\mathbf{i}^\mathrm{T}\mathbf{x}}$ defines a sign, and $\mathbf{i} \in \{0,1\}^n$ is the binary representation of index $i$.  \par

Any pseudo-boolean function can be represented using Walsh coefficients. Each coefficient is associated with a mask that marks variables. If at least one mask marks a given pair of variables, then they are non-linearly dependent. Using Walsh coefficients, we can compute a function value for any solution. To this end, we must compute signs of Walsh coefficients by summarizing the binary values of variables marked by a given mask. If the sum is even, then the Walsh coefficient is added or subtracted otherwise. For instance, for a 3-bit function represented by two Walsh coefficients $w_{111}=-5$ and $w_{001}=2$, the function value for solution $101$ will be: $w_{111} - w_{001} = -5-2=-7$.\par

Consider $f_{px}(x_1,...,x_4)=xor(x_1,x_2)+xor(x_2,x_3) + xor(x_3,x_4)$. Note that $xor$ introduces non-linear dependencies between variables. Thus, Walsh decomposition will identify pairs $(x_1,x_2)$, $(x_2,x_3)$, and $(x_3,x_4)$ as dependent. The above representation is the additive form of $f_{px}$, in which subfunctions (each $xor$) share variables, i.e., overlap. If subfunctions overlap, the function may be hard to optimize because by modifying a shared variable, we modify the output of more than one subfunction at a time. Information about variable dependencies is stored in the Variable Interaction Graph (VIG), which can be represented by a square matrix denoting whether a given pair of variables is dependent or not. VIG for $f_{px}$ is presented in Table S-III (supplementary material). Using VIG, we can propose masks that improve the performance of variation operators. Partition crossover (PX) \cite{pxForBinary} is a VIG-based mixing operator. For two individuals, it removes all VIG entries that refer to variables equal in mixed individuals and clusters variables based on the remaining ones. For individuals $\vec{x_{p1}}=1110$ and $\vec{x_{p2}}=0011$ it will create two mixing masks $m_{px1}=\{1,2\}$ and $m_{px2}=\{4\}$. An important feature of PX is that even if subfunctions overlap, it creates masks that allow exchanging subfunction arguments without breaking them, i.e., all the subfunction argument sets in the offspring will be the same as in one of the parents. If we choose $m_{px1}$ and copy genes from $\vec{x_{p2}}$ to $\vec{x_{p1}}$, then we will receive $\vec{x_{p3}}=0010$ (first $xor$ has the same arguments as $\vec{x_{p1}}$, second is the same for both parents, third is the same as in $\vec{x_{p2}}$). Therefore, PX is useful in optimizing overlapping problems \cite{dgga}. However, if the epistasis of a VIG is high, i.e., all or almost all variables are considered dependent, then PX may create a single mask marking all differing genes. Since such mask is useless, identifying and ignoring dependencies irrelevant to the optimization process can increase effectiveness \cite{superMasksGray}. \par

\subsection{Linkage trees and SLL-using optimizers}
\label{sec:relWork:lts}
In black-box optimization, variable dependencies can be discovered. Statistical Linkage Learning (SLL) \cite{P3Original,dsmga2,ltga} predicts the eventual dependencies by analyzing the entropy of variable value frequencies that occur in the population. The Dependency Structure Matrix (DSM) stores the predicted dependency strength between variable pairs. In SLL, we do not know which dependencies are true. Therefore, many SLL-using optimizers cluster DSM to create Linkage Trees (LTs). In an LT, leaves refer to single variables. The most dependent nodes (concerning DSM) are joined and form larger clusters. LT root is a full mask and is useless for optimization. However, the other LT nodes can be used as variation masks.\par

LT-GOMEA is a state-of-the-art SLL-using optimizer \cite{ltgaPopulationSizing}. During its run, it creates subpopulations of increasing size. In each iteration of a given subpopulation, LT-GOMEA builds DSM and LT. It uses Optimal Mixing (OM) to mix individuals. OM involves the \textit{source} individual that is being updated, the \textit{donor} individual and a mask. Genes marked by a mask are copied from the donor to the source. If this operation does not decrease the source fitness, its result is preserved or rejected otherwise. During OM, all LT masks are considered, and a different donor is randomly chosen for each mask.\par

%Parameter-less Population Pyramid (P3) uses SLL and OM, but it also introduces a new way of population management \cite{P3Original}. P3 maintains many subpopulations that resemble a pyramid and are called \textit{levels}. In each iteration, a new individual (\textit{climber}) is created. First, it is improved using the First Improvement Hillclimber (FIHC). FIHC flips bits in a random order. If such an operation improves fitness, it is preserved or rejected otherwise. After FIHC a copy of the climber is added to the first level of the pyramid. Then, the climber is mixed using OM of subsequent pyramid levels. If OM improves it, then an improved climber's copy is added to the next pyramid level (it is created if it does not exist yet). P3 preserves population diversity well and uses a separate LT for each pyramid level. These features make it effective in solving overlapping problems \cite{3lo}.

Parameter-less Population Pyramid (P3) uses SLL and OM, but it also introduces a new way of population management \cite{P3Original}. P3 maintains many subpopulations that resemble a pyramid and are called \textit{levels}. In each iteration, P3 creates a new individual (\textit{climber}). First, it is improved using the First Improvement Hillclimber (FIHC). FIHC flips bits in a random order. If a bitflip improves fitness, it is preserved or rejected otherwise. After FIHC, a copy of the climber is added to the first level of the pyramid. Then, the climber is mixed using OM of subsequent pyramid levels. If OM improves it, then an improved climber's copy is added to the next pyramid level (it is created if it does not exist yet). P3 preserves population diversity well and uses a separate LT for each pyramid level. These features make it effective in solving overlapping problems \cite{3lo}.

\subsection{Dependency types}
\label{sec:relWork:deps}
If gray-box mechanisms are unavailable, instead of using SLL, we can discover variable dependencies using variable dependency checks. In non-linearity check \cite{linc} $x_g$ and $x_h$ are dependent if: $f(\vec{x}) + f(\vec{x}^{g,h}) \neq f(\vec{x}^g) + f(\vec{x}^h)$, where $\vec{x}^{g}$, $\vec{x}^h$, and $\vec{x}^{g,h}$ are the individuals obtained from $\vec{x}$ by flipping $x_g$, $x_h$ or both of them.\par
    %\begin{equation}
    %\small
    %\label{eq:nonLinear}
    %    f(\vec{x}) + f(\vec{x}^{g,h}) \neq f(\vec{x}^g) + f(\vec{x}^h)
    %\end{equation}
    %where $\vec{x}^{g}$, $\vec{x}^h$, and $\vec{x}^{g,h}$ are the individuals obtained $\vec{x}$ by flipping $x_g$, $x_h$ or both of them, respectively.\par

    In some cases, the non-linearity check may discover dependencies that are irrelevant to the optimization process. Such dependencies should be ignored or eliminated because they can deteriorate the quality of variation masks \cite{GoldMonotonicity}. To this end, the non-monotonicity check can be used \cite{FIHCwLL,2dled}, which finds $x_g$ and $x_h$ dependent if at least one of the following conditions is true:\\
    \footnotesize
    \textbf{C1.} $f(\vec{x}) < f(\vec{x}^g)  \land f(\vec{x}^h) \geq f(\vec{x}^{g,h})$   \textbf{C4.} $f(\vec{x}) < f(\vec{x}^h)  \land f(\vec{x}^g) \geq f(\vec{x}^{g,h})$\\
    \textbf{C2.} $f(\vec{x}) = f(\vec{x}^g)  \land f(\vec{x}^h) \neq f(\vec{x}^{g,h})$    \textbf{C5.} $f(\vec{x}) = f(\vec{x}^h)  \land f(\vec{x}^g) \neq f(\vec{x}^{g,h})$\\
    \textbf{C3.} $f(\vec{x}) > f(\vec{x}^g)  \land f(\vec{x}^h) \leq f(\vec{x}^{g,h})$   \textbf{C6.} $f(\vec{x}) > f(\vec{x}^h)  \land f(\vec{x}^g) \leq f(\vec{x}^{g,h})$\\
    %C4. $f(\vec{x}) < f(\vec{x}^h)  \land f(\vec{x}^g) \geq f(\vec{x}^{g,h})$\\
    %C5. $f(\vec{x}) = f(\vec{x}^h)  \land f(\vec{x}^g) \neq f(\vec{x}^{g,h})$\\
    %C6. $f(\vec{x}) > f(\vec{x}^h)  \land f(\vec{x}^g) \leq f(\vec{x}^{g,h})$\\
\normalsize
Directional Direct Linkage Discovery (2DLED) \cite{2dled} reduces irrelevant dependencies by considering non-symmetrical dependencies. Using C1-C6 conditions, it finds $x_g$ dependent on $x_h$ if C1-C3 holds, and $x_h$ dependent on $x_g$ if C4-C6 holds. It can discover symmetrical or non-symmetrical dependency between two variables. 

%For some problems, it may be crucial to construct high-quality variation masks \cite{2dled}.

\section{PX with Weighted Dynamic VIG}
\label{sec:PXwdVIG}

\subsection{Intuitions and motivations}
\label{sec:PXwdVIG:intutions}
Let us consider a 6-bit problem, which Walsh decomposition is built from four coefficients: $w_{110000} = w_{011000} = w_{001100} = w_{111111} = 1$. We wish to mix two individuals: 
\begin{itemize}
    \item $\vec{x_a}=[111101] = w_{110000} + w_{011000} + w_{001100} - w_{111111} = 2$
    \item $\vec{x_b}=[100010] = -w_{110000} + w_{011000} + w_{001100} + w_{111111} = 2$
\end{itemize}
We may say, we wish to pass the positive sign of $w_{110000}$ from $\vec{x_a}$ to $\vec{x_b}$ or we wish to pass the positive sign of $w_{111111}$ from $\vec{x_b}$ to $\vec{x_a}$. However, for the considered problem, the Walsh-based VIG will be a full graph. Thus, any PX mask will mark all genes and will be useless. The question is whether we must cluster all genes together in this case. Below we analyze two available scenarios.\par

In the first scenario, we wish to improve $\vec{x_b}$. Thus, we have to copy $x_2$ from $\vec{x_a}$ to pass the positive value of $w_{110000}$, which will change the sign of $w_{011000}$ in the modified $\vec{x_b}$. Therefore, we have to copy $x_3$ too, which will change the sign of $w_{001100}$, and, therefore, we have to copy $x_4$. Note that that whenever we flip $x_2$ to modify the sign of $w_{01100}$, we have to flip one more variable joined with $x_2$ by the masks of other Waslh coefficients $x_2$ belongs to. Note that all dependent gene pairs which dependence arise from the masks of size 2 can be considered unbreakable, i.e., if we want to preserve the sign of such Walsh coefficient and flip one of its variables, then we must flip the other variable too.\par

Copying $x_2$, $x_3$, and $x_4$ from $\vec{x_a}$ to $\vec{x_b}$ will create the following solution $\vec{x_b}'=[111110]$, in which the value of $w_{111111}$ will be negative. Therefore, after copying $x_4$, we have to copy \textbf{\textit{one}} more gene, $x_5$ or $x_6$, i.e., we can choose one of the remaining genes marked by $w_{111111}$. Thus, we do not have to use a mask that covers all dependencies arising from $w_{111111}$. \par

In the second scenario, we wish to improve $\vec{x_a}$ by passing the positive value of $w_{11111}$, but do not modify the genes marked by $w_{110000}$, $w_{011000}$, and $w_{001100}$. Therefore, we have to copy $x_5$ or $x_6$ from $\vec{x_b}$ to $\vec{x_a}$. Although $w_{111111}$ marks six genes, we must copy only one to pass the positive value of $w_{111111}$.\par

The above examples show that gene relations arising from long masks seem much weaker than those arising from short masks. Let us consider an example of a 7-bit problem, which Walsh decomposition is built from four coefficients: $w_{1100000}=w_{0111100}=w_{0000110}=w_{0000011}=1$. We wish to mix two individuals:\\
$\vec{x_c}=[1110011] = w_{1100000} + w_{0111100} - w_{0000110} + w_{0000011} = 2$\\
$\vec{x_d}=[0010110] = w_{1100000} + w_{0111100} + w_{0000110} - w_{0000011} = 2$\\
We wish to pass the positive value of $w_{0000110}$ from $\vec{x_d}$ to $\vec{x_c}$. Therefore, we copy $x_5$ from $\vec{x_d}$ to $\vec{x_c}$, which makes $w_{0111100}$ negative. The mask size of $w_{0111100}$ is 4 but except $x_5$ the only gene covered by $w_{0111100}$ and differing in $\vec{x_c}$ and $\vec{x_d}$ is $x_2$. Thus, when $\vec{x_c}$ and $\vec{x_d}$ are mixed, mask $w_{0111100}$ induces the dependency only between $x_2$ and $x_4$ and acts like the mask of size 2. Therefore, we copy $x_2$, then, we copy $x_1$ and obtain $\vec{x_c}'=[0010111] = w_{1100000} + w_{0111100} + w_{0000110} + w_{0000011} = 4$.\par

To recap, when mixing two individuals, the dependency strength raised by the Walsh coefficient mask depends on the number of differing positions it covers, not on its pure size. For a given pair of individuals, a large mask indicates strong dependencies if the number of differing positions it covers is close or equal to two (lower does not indicate any dependency). Consequently, if a large mask covers many differing positions, its influence on the dependency strength between the covered variables seems less significant.

\subsection{Weighted VIG propositions}
\label{sec:PXwdVIG:wVIG}

Using the intuitions explained above, we define the Weighted Dynamic VIG (wdVIG). wdVIG is computed for a given pair of mixed solutions. The value of the wdVIG cell is defined as follows.

\begin{equation} 
\small
    \label{eq:wdVIG}
    \begin{aligned}
        wdVIG(\vec{x_a}, \vec{x_b}, g, h) = &\sum_{mask}^{masks(\vec{x_a},\vec{x_b},g,h)}      
                |w_{mask}|\cdot\\
                &\cdot \frac{2}{d(mask,\vec{x_a},\vec{x_b}) \bigl(d(mask,\vec{x_a},\vec{x_b})-1\bigr)} 
    \end{aligned}
\end{equation}
where $\vec{x_a}$ and $\vec{x_b}$ are the mixed individuals, $g$ and $h$ are the genes for which we compute the strength of dependency, $masks(\vec{x_a},\vec{x_b},g,h)$ is the set of all non-zero Walsh coefficients, which masks cover $x_g$ and $x_h$ and for which $x_g$ and $x_h$ differ in $\vec{x_a}$ and $\vec{x_b}$, $d(mask,\vec{x_a},\vec{x_b})$ is the number of genes differing in $\vec{x_a}$ and $\vec{x_b}$ covered by the $mask$. Thus, the absolute value of the coefficient is divided by the number of dependencies it raises for a given pair of individuals.\par

%Consider the 4-bit problem, which Walsh decomposition is built from two coefficients: $w_{1110}=w_{0111}=1$ and two mixed individuals: $\vec{x_e}=[0001]$ and $\vec{x_f}=[1111]$. In wdVIG, The dependency for $x_2$ and $x_3$ will be: $wdVIG(\vec{x_e},\vec{x_f}, 2, 3) = |w_{1110}|\cdot\frac{2}{3(3-1)} + |w_{0111}|\cdot\frac{2}{2(2-1)} = \frac{|w_{1110}|}{3}+|w_{0111}|$. Coefficient $w_{1110}$ raises three dependency pairs for $\vec{x_e}$ and $\vec{x_f}$, i.e. (1,2), (1,3), and (2,3). Therefore, its influence (the absolute value of the coefficient) is divided by three. Situation differs for $w_{0111}$, which for individuals $\vec{x_e}$ and $\vec{x_f}$ raises only one dependency pair, between $x_2$ and $x_3$, because $x_4$ is equal in both individuals. Note that for $\vec{x_e}$ and $\vec{x_f}$, all dependencies involving $x_4$ equal 0 because the value of $x_4$ is the same for both individuals.\par

Consider the 4-bit problem, which Walsh decomposition is built from two coefficients: $w_{1110}=w_{0111}=1$ and two mixed individuals: $\vec{x_e}=[0001]$ and $\vec{x_f}=[1111]$. In wdVIG, The dependency for $x_2$ and $x_3$ will be: $wdVIG(\vec{x_e},\vec{x_f}, 2, 3) = |w_{1110}|\cdot\frac{2}{3(3-1)} + |w_{0111}|\cdot\frac{2}{2(2-1)} = \frac{|w_{1110}|}{3}+|w_{0111}|$. Coefficient $w_{1110}$ raises three dependency pairs for $\vec{x_e}$ and $\vec{x_f}$, i.e. (1,2), (1,3), and (2,3). Therefore, its influence (the absolute value of the coefficient) is divided by three. Oppositely, for $\vec{x_e}$ and $\vec{x_f}$, $w_{0111}$ raises only one dependency pair (2,3) because $x_4$ is equal $\vec{x_e}$ and $\vec{x_f}$.\par

In wdVIG, the dependency strength depends on the absolute coefficient value and the number of dependencies raised by the coefficient's mask for a given individual pair. To verify wdVIG quality, we will consider three other weighted VIG types.

%The wdVIG values can be different for different pairs of mixed individuals. We define the dependency strength using the absolute value of a coefficient and the number of dependencies raised by the coefficient's mask. To experimentally verify if our intuitions are valid, we will consider three other types of weighted VIG.

\begin{enumerate}
    \item wdVIG without mask size influence (wdVIGns) defined as 

    \begin{equation}
    \small
    \label{eq:wdVIGnsize}
        wdVIGns(\vec{x_a}, \vec{x_b}, g, h) = \sum_{mask}^{masks(\vec{x_a},\vec{x_b},g,h)}  |w_{mask}|
    \end{equation}
    In wdVIGns, the influence of each coefficient (covering $x_g$ and $x_h$) equals zero if $x_g$ or $x_h$ are equal in the mixed individuals. Otherwise, it equals the absolute coefficient value.
    %For wdVIGns, the influence of each coefficient (whose mask covers $x_g$ and $x_h$) on the dependency between $x_g$ and $x_h$ is equal to the coefficient value. However, if $x_g$ or $x_h$ are equal in the mixed individuals, then the value of the wdVIGns entry equals zero.

    \item weighted static VIG (wsVIG) that equals wdVIG but ignores genotype differences between mixed individuals, defined as

    \begin{equation}
    \small
    \label{eq:wsVIG}
        wsVIG(g, h) = \sum_{mask}^{masks(g,h)}      
                |w_{mask}|\cdot \frac{2}{size(mask) \bigl(size(mask)-1\bigr)} 
    \end{equation}
    where $masks(g,h)$ is the set of all Walsh coefficients whose masks mark $x_g$ and $x_h$, and $size(mask)$ is the number of genes marked by the whole mask.\par

    %The only difference between wdVIG and wsVIG is that the latter does not take into account the differences between mixed individuals, i.e., wsVIG is equal to wdVIG when mixed individuals are complementary.

    \item wsVIG without mask size influence (wsVIGns) considers only the absolute values of all coefficients that mark a given pair of genes. wsVIGns is defined as follows.

    \begin{equation}
    \small
    \label{eq:wsVIGnsize}
        wsVIGns(g, h) = \sum_{mask}^{masks(g,h)} |w_{mask}| 
    \end{equation}
    
\end{enumerate}

 \begin{table*}  
    \caption{Weighted VIGs for a 6-bit problem built from four coefficients ($w_{111000} = 10$, $w_{110101} = 8$, $w_{000111} = 7$, and $w_{010100} = 2$) and individuals $\vec{x_o}=[101000]$, $\vec{x_p}=[010110]$.} 
    \label{tab:PXwdVIG:wVIG}
    \scriptsize
    \begin{subtable}{.24\textwidth}
    \centering 
       \begin{tabular}{l|cccccc}
              & 1  & 2  & 3  & 4  & 5 & 6 \\
              \hline
            1 & X  & 18 & 10 & 8  & 0 & 8  \\
            2 & 18 & X  & 10 & 10 & 0 & 8  \\
            3 & 10 & 10 & X  & 0  & 0 & 0  \\
            4 & 8  & 10 & 0  & X  & 7 & 15 \\
            5 & 0  & 0  & 0  & 7  & X & 7  \\
            6 & 8  & 8  & 0  & 15 & 7 & X 
        \end{tabular}
    \caption{wsVIGns}
    \label{tab:PXwdVIG:wVIG:wsVIGnsize}
    \end{subtable}
    \begin{subtable}{.24\textwidth}
    \centering 
       \begin{tabular}{l|cccccc}
         & 1 & 2 & 3 & 4 & 5 & 6 \\
          \hline
            1 & X   & 4.7 & 3.3 & 1.3 & 0.0 & 1.3 \\
            2 & 4.7 & X   & 3.3 & 3.3 & 0.0 & 1.3 \\
            3 & 3.3 & 3.3 & X   & 0.0 & 0.0 & 0.0 \\
            4 & 1.3 & 3.3 & 0.0 & X   & 2.3 & 3.7 \\
            5 & 0.0 & 0.0 & 0.0 & 2.3 & X   & 2.3 \\
            6 & 1.3 & 1.3 & 0.0 & 3.7 & 2.3 & X     
        \end{tabular}
    \caption{wsVIG}
    \label{tab:PXwdVIG:wVIG:wsVIG}
    \end{subtable}
    \begin{subtable}{.24\textwidth}
    \centering 
       \begin{tabular}{l|cccccc}
         & 1 & 2 & 3 & 4 & 5 & 6 \\
          \hline
            1 & X  & 18 & 10 & 8  & 0 & 0 \\
            2 & 18 & X  & 10 & 10 & 0 & 0 \\
            3 & 10 & 10 & X  & 0  & 0 & 0 \\
            4 & 8  & 10 & 0  & X  & 7 & 0 \\
            5 & 0  & 0  & 0  & 7  & X & 0 \\
            6 & 0  & 0  & 0  & 0  & 0 & X
        \end{tabular}
    \caption{wdVIGns}
    \label{tab:PXwdVIG:wVIG:wdVIGnsize}
    \end{subtable}
    \begin{subtable}{.24\textwidth}
    \centering 
       \begin{tabular}{l|cccccc}
         & 1 & 2 & 3 & 4 & 5 & 6 \\
          \hline
            1 & X   & 6.0 & 3.3 & 2.7 & 0.0 & 0.0 \\
            2 & 6.0 & X   & 3.3 & 4.7 & 0.0 & 0.0 \\
            3 & 3.3 & 3.3 & X   & 0.0 & 0.0 & 0.0 \\
            4 & 2.7 & 4.7 & 0.0 & X   & 7.0 & 0.0 \\
            5 & 0.0 & 0.0 & 0.0 & 7.0 & X   & 0.0 \\
            6 & 0.0 & 0.0 & 0.0 & 0.0 & 0.0 & X 
        \end{tabular}
    \caption{wdVIG}
    \label{tab:PXwdVIG:wVIG:wdVIG}
    \end{subtable}
\end{table*}

\begin{table}[ht]
\caption{Individuals $\vec{x_o}=[101000]$, $\vec{x_p}=[010110]$ and the signs of the Walsh coefficients.} 
    \label{tab:wdVIGmixExample}
\scriptsize
\begin{tabular}{l|cccccc|cccc}
       & $x_1$ & $x_2$ & $x_3$ & $x_4$ & $x_5$ & $x_6$ & $w_{111000}$   & $w_{110101}$   & $w_{000111}$   & $w_{010100}$   \\
       \hline
$x_o$   & 1 & 0 & 1 & 0 & 0 & 0 & {\small\texttt{+}} & {\small\texttt{-}} & {\small\texttt{+}} & {\small\texttt{+}} \\
$x_p$   & 0 & 1 & 0 & 1 & 1 & 0 & {\small\texttt{-}} & {\small\texttt{+}} & {\small\texttt{+}} & {\small\texttt{+}} \\
\textbf{diff}   & 1 & 1 & 1 & 1 & 1 & * &  &  &  &  \\
$x_r$ & 0 & 1 & 1 & 1 & 1 & 0 & {\small\texttt{+}} & {\small\texttt{+}} & {\small\texttt{+}} & {\small\texttt{+}}
\end{tabular}
\end{table}

We use LTs to cluster the four weighted VIG types proposed above. To show the differences between them, let us consider a 6-bit problem built from four coefficients ($w_{111000} = 10$, $w_{110101} = 8$, $w_{000111} = 7$, and $w_{010100} = 2$). We wish to improve $\vec{x_o}=[101000]$ by mixing with  $\vec{x_p}=[010110]$. In Table \ref{tab:PXwdVIG:wVIG}, we present all four considered weighted VIG types for $\vec{x_o}$ and $\vec{x_p}$. In Table \ref{tab:wdVIGmixExample}, we present the signs of Walsh coefficients related to $\vec{x_o}$ and $\vec{x_p}$. Table \ref{tab:wdVIGmixExample} shows that to improve $\vec{x_o}$ by mixing with $\vec{x_p}$, we need to pass the positive sign of $w_{110101}$ and preserve the signs of other coefficients. Thus, we have to copy gene $x_4$ or genes $x_1$, $x_2$, and $x_4$ together (because $x_6$ is the same in $\vec{x_o}$ and $\vec{x_p}$). We do not want to copy $x_3$ in both cases. Starting from $x_4$, we have to copy it with $x_2$ and $x_5$ to preserve the signs of $w_{010100}$ and $w_{000111}$, respectively. If we copy $x_2$ and $x_4$ together, then we have to copy $x_1$ to pass the positive sign of $w_{110101}$. Thus, we wish to use the following mask $\{1,2,4,5\}$. If we always choose the strongest relation to join LT nodes, then obtaining such a mask for VIGs presented in Tables \ref{tab:PXwdVIG:wVIG:wsVIGnsize}-\ref{tab:PXwdVIG:wVIG:wdVIGnsize} is impossible because they will first join $x_1$ and $x_2$. We ignore the relation between $x_6$ and $x_4$ because $x_6$ is equal in both individuals. Then, the strongest relation will be between pairs ($x_1$,$x_3$),  ($x_2$,$x_3$), and ($x_2$,$x_4$). Thus, $x_3$ will be added to the node ($x_1$,$x_2$). Such a mask will force passing the sign of $w_{111000}$ from $x_p$ to $x_o$, which will deteriorate $x_o$ quality.\par

The situation differs for wdVIG. First, we will join ($x_4$,$x_5$), then ($x_1$,$x_2$) and the strongest relation between $x_4$ and $x_2$ will join these two nodes into ($x_1$,$x_2$,$x_4$,$x_5$), which is the mask we wish to obtain. 

\subsection{Weighted PX}
\label{sec:PXwdVIG:wPX}

For problems with high epistasis, PX may become ineffective due to generating masks covering all differences in the mixed individuals (see Section \ref{sec:relWork:grayWalsh}). Therefore, we propose Weighted PX (wPX), which does not suffer from this disadvantage. In wPX, we use weighted VIG and cluster it using LT. LTs are frequently used as part of OM in SLL-using black-box-dedicated optimizers (see Section \ref{sec:relWork:lts}). 

%In P3 \cite{P3Original}, LT nodes (masks) of the same size are considered in a random order, but the shorter masks go first. In LT-GOMEA \cite{ltgaPopulationSizing}, all masks are considered in a random order, and the mask size does not influence it. \par

When we use SLL in black-box optimization, we do not know which dependencies are direct. Moreover, we do not even know which are true or false. The relation strength for each gene pair is a result of statically-based prediction. Therefore, the strategy to consider all LT nodes (or almost all, since sometimes the nodes with size $1$ are ignored) seems justified. However, in gray-box optimization, the situation differs. We know the true dependencies and (at least in the research presented in this work) the Walsh coefficients from which these dependencies arise. Thus, we can choose only those LT nodes that are worth consideration.\par

%\begin{figure}[h]
%	\centering
%	\includegraphics[width=0.6 \linewidth]{images/wPXtreeUp_v2.png}
%	\caption{An example of LT node selection made using the \textit{LTtop} strategy for a 12-bit problem. Green and blue nodes are used for mixing (green go first). The blue nodes are equivalent to standard PX masks.}
%	\label{fig:LTtop}
%\end{figure}

In wPX, we assume that we obtain masks from the weighted VIG in which those gene pairs that are the arguments of subfunctions with the highest influence on fitness are highly dependent. Oppositely, the relation strength of those gene pairs that are the arguments of subfunctions with the low influence on fitness is low. In LT created on the base of such weighted VIG, the nodes on the lowest levels of LT will consist of those variables that are strongly related, while the weakest relations will cause the creation of nodes on top of the tree. If, for a given weighted VIG, a standard PX would create more than one mixing mask, then the same number of disjoint LTs will be created. We propose the following strategy of choosing LT nodes for wPX (denoted as \textit{LTtop}). In LTtop, we consider LT roots if more than one LT was created, as well as all nodes that are one level beneath the roots. We ignore all nodes that are full masks (if a single LT was created) and of size 1. Thus, the LTtop strategy will limit the number of considered mixing masks, which shall spare the computation resources. At the same time, it will tend to break the weakest relations that caused the creation of nodes right beneath the LT root (for the example, see Section S-V and Fig. S-II, supplementary material).\par

%An example of the LTtop node selection is presented in Fig. \ref{fig:LTtop}. We select all green and blue nodes. We consider both LT roots (blue nodes are equivalent to PX masks) because none of them is a full mask and ignore node \{12\} because it marks only one gene. All nodes are considered in random order, but before considering any LT root, the selected nodes connected to it must be considered first. For instance, in the example presented in Figure \ref{fig:LTtop}, we can consider node \{9,10,11,12\} before node \{5,6,7,8\} if node \{9,10,11\} was considered before \{9,10,11,12\}.\par

Similarly to Optimal Mixing employed in P3 and LT-GOMEA \cite{P3Original,ltgaPopulationSizing}, in wPX, we try to improve a single individual denoted as \textit{source}. To this end, we randomly choose the \textit{donor} individual, generate weighted VIG for the considered individual pair, obtain LT, and use LTtop-selected nodes as mixing masks. If copying genes from the donor to the source individual does not improve the source individual, then the modification to its genotype is rejected. wPX ends if the source individual is improved or none of the considered masks were used. \par

\subsection{Gray-box Optimizer for Problems with High Epistasis}
\label{sec:PXwdVIG:GBOphe}

To utilize wPX, we propose Gray-box Optimizer for Problems with High Epistasis
(GBO-PHE) that refers to the P3-like population management. GBO-PHE is parameter-less, which seems useful for practical purposes. Its procedure is presented in Pseudocode \ref{alg:PXwdVIG:GBOphe}.

\begin{algorithm}
	\caption{The general procedure of GBO-PHE}
    \begin{flushleft}
	\begin{algorithmic}[1]
        \State $Pyramid \leftarrow $ empty; \label{line:PXwdVIG:GBOphe:initPyr}
        \State $WalshCoeffs \leftarrow$ ComputeWalshCoeffs();\label{line:PXwdVIG:GBOphe:coeffs}
        \While {$\neg StopCondition$}
            \State $\vec{x} \gets$ FIHC(CreateRandomInd()); \label{line:PXwdVIG:GBOphe:createClimber}
            %\State $\vec{x} \gets$ FIHC($\vec{x}$);\label{line:PXwdVIG:GBOphe:fihc}
            \State AddToPyramidLevel($Pyramid$,$0$,$\vec{x}$); \label{line:PXwdVIG:GBOphe:level0}
            \For{\textbf{each} $level$ \textbf{in} $Pyramid$}
                \State $fitnessOld \gets$ Fitness($\vec{x}$); 
                \For{\textbf{each} $ind$ \textbf{in (random order)} $level$} \label{line:PXwdVIG:GBOphe:wPXomStart}
                    \State $\vec{x} \gets$ wPX($\vec{x}$, $ind$, $WalshCoeffs$); \label{line:PXwdVIG:GBOphe:wPXomEnd}
                \EndFor
                \If {Fitness($\vec{x}$) $> fitnessOld$}
                    \State AddToPyramidLevel($Pyramid$,$level+1$,$\vec{x}$); \label{line:PXwdVIG:GBOphe:levelPlus}
                \EndIf
            \EndFor
        \EndWhile
     
	\end{algorithmic}
    \end{flushleft}
	\label{alg:PXwdVIG:GBOphe}
\end{algorithm}

GBO-PHE starts from initializing the pyramid and computing Walsh coefficients (lines \ref{line:PXwdVIG:GBOphe:initPyr}-\ref{line:PXwdVIG:GBOphe:coeffs}). In each iteration, a new individual $\vec{x}$ (\textit{climber} in P3) is created randomly, optimized by FIHC (line \ref{line:PXwdVIG:GBOphe:createClimber}), and added to the lowest level of the pyramid (line \ref{line:PXwdVIG:GBOphe:level0}). Then, $\vec{x}$ is mixed with the subsequent pyramid levels. While mixed with each level, $\vec{x}$ acts as the source individual of wPX, while all individuals at a given level are considered in the random order and act as donors (lines \ref{line:PXwdVIG:GBOphe:wPXomStart}-\ref{line:PXwdVIG:GBOphe:wPXomEnd}). If mixing with a given level improves the climber's fitness, then its improved copy is added to the next pyramid level (if necessary, a new level is created and initialized with the improved copy of $\vec{x}$).\par

%(if such a level does not exist, it is created and initialized with the improved copy of $\vec{x}$).\par

\section{Revealing Hidden Problem Structure}
\label{sec:expInit}
In this work, we measure the dependency strength between variables to filter and use only the strongest dependencies that are the most relevant to the optimization process. In this section, we present the results that justify such a choice and aim to answer the following research questions:\\
\textbf{RQ1.} If we add noise to the problem of a known structure, then how will this influence the Walsh-based problem representation and the VIGs considering non-linear, non-monotonical, and 2DLED dependencies?\\
\textbf{RQ2.} Can we filter the noise using Walsh coefficients?\\
\textbf{RQ3.}  Can we filter the noise in real-world problems' instances and obtain a strong problem structure for problems that refer to fully connected VIGs?

\subsection{Noise modelling and filtering}
\label{sec:expInit:noise}
%To investigate the influence of noise on the problem structure, we propose the following experiment based on the \textit{Onemax} problem defined as follows.
%\begin{equation}
%    \label{eq:onemax}
%    \mathit{onemax(u)}=u
%\end{equation}
%where $u$ is the sum of gene values (so called \textit{unitation}).\par

To investigate the influence of noise on the problem structure, we propose the experiment using the \textit{onemax} problem defined as $\mathit{onemax(\vec{x})}=u(\vec{x})$, where $u(\vec{x})$ is the sum of gene values (\textit{unitation}). We consider 10-bit onemax instances. To the value of each problem solution, we add a random real value from the range $(0, nVol)$ where $nVol$ is the maximum noise volume. The problem is \textit{static}, i.e., the noise distorting each solution is chosen once. Thus, each solution always evaluates to the same function value. The considered instances are of toy size because we wish to investigate their complete Walsh representation, i.e., we must compute all Walsh coefficients of all possible sizes. For each problem instance, we construct VIG based on the non-linearity, non-monotonicity, and 2DLED checks. Since the problem instances are of toy size, we perform each check in every available context. In onemax, all variables are independent. Thus, all obtained dependencies will originate from noise.\par

\begin{figure}[ht]
    \begin{subfigure}[b]{0.49\linewidth}
		\resizebox{\linewidth}{!}{%
			\tikzset{every mark/.append style={scale=2.5}}
			\begin{tikzpicture}
			\begin{axis}[%
            legend entries={non-linearity, DLED, 2DLED},
            legend columns=-1,
            legend to name=named,
			%legend columns=-1,
            %legend entries={dgGA,cGOMEA,P3,P3-DLED,LT-GOMEA-DLED},
            %legend to name=named,
            %legend columns=-1,
            %legend entries={LT-GOMEA-DLED,LT-GOMEA-SLL,P3-DLED,P3-SLL},
            %legend to name=named,
			xtick={1,2,3,4,5,6,7,8,9,10,11,12},
			xticklabels={1.0, ,1.2, ,1.4, ,1.6, , 1.8, , 2.0, 2.5},
			xmin=1,
			xmax=12,
			%ymode=log,
			ymin=0,
			ymax=1,
			%legend pos=south east,
			xlabel=\textbf{maximum noise volume ($nVol$)},
			ylabel=\textbf{VIG epistasis (VIG fill)},
			grid,
			grid style=dashed,
			ticklabel style={scale=1.5},
			label style={scale=1.5},
			legend style={font=\fontsize{8}{0}\selectfont}
			]

            %\addplot[
			%color=brown,
			%mark=triangle*,
			%]
			%coordinates {
			%	(1,477026)(2,1961269)(3,7857276.5)(4,11729329)(5,15856165)(6,30264820)(7,24703945)(8,32664404)
			%};
   
            %\addplot[
			%color=black,
			%mark=*,
			%]
			%coordinates {
            %    (1,439884.5)(2,1464475.5)(3,4787594)(4,7602447)(5,13019899.5)(6,22470466)(7,27529834)(8,37681398)
			%};

            \addplot[
			color=blue,
            name path=nonLinMax,
            forget plot
			]
			coordinates {
				(1,1)(2,1)(3,1)(4,1)(5,1)(6,1)(7,1)(8,1)(9,1)(10,1)(11,1)(12,1)
			};
   
            \addplot[
			color=blue,
			mark=*,
            name path=nonLinMed
			]
			coordinates {
				(1,1)(2,1)(3,1)(4,1)(5,1)(6,1)(7,1)(8,1)(9,1)(10,1)(11,1)(12,1)
			};

            \addplot[
			color=blue,
            name path=nonLinMin,
            forget plot
			]
			coordinates {
				(1,1)(2,1)(3,1)(4,1)(5,1)(6,1)(7,1)(8,1)(9,1)(10,1)(11,1)(12,1)
			};

            \addplot[
			color=violet,
            name path=dledMax,
            forget plot
			]
			coordinates {
				(1,0)(2,0.53333333)(3,0.06666667)(4,0.57777778)(5,0.44444444)(6,0.37777778)(7,0.77777778)(8,0.71111111)(9,0.71111111)(10,0.8)(11,0.8)(12,0.71111111)

			};

            \addplot[
			color=violet,
			mark=triangle*,
            name path=dledMed
			]
			coordinates {
				(1,0)(2,0)(3,0)(4,0.055555555)(5,0.06666667)(6,0.1)(7,0.18888889)(8,0.22222222)(9,0.22222222)(10,0.22222222)(11,0.255555555)(12,0.444444445)

			};

            \addplot[
			color=violet,
            name path=dledMin,
            forget plot
			]
			coordinates {
				(1,0)(2,0)(3,0)(4,0)(5,0)(6,0)(7,0)(8,0)(9,0.04444444)(10,0.06666667)(11,0.04444444)(12,0.13333333)
			};

            \addplot[
			color=red,
            name path=2dledMax,
            forget plot
			]
			coordinates {
				(1,0)(2,0.3)(3,0.04444444)(4,0.32222222)(5,0.23333333)(6,0.2)(7,0.56666667)(8,0.46666667)(9,0.46666667)(10,0.51111111)(11,0.54444444)(12,0.47777778)

			};

            \addplot[
			color=red,
			mark=square*,
            name path=2dledMed
			]
			coordinates {
				(1,0)(2,0)(3,0)(4,0.03333333)(5,0.038888885)(6,0.05555556)(7,0.1)(8,0.127777775)(9,0.127777775)(10,0.127777775)(11,0.16666667)(12,0.266666665)

			};

            \addplot[
			color=red,
            name path=2dledMin,
            forget plot
			]
			coordinates {
				(1,0)(2,0)(3,0)(4,0)(5,0)(6,0)(7,0)(8,0)(9,0.02222222)(10,0.03333333)(11,0.02222222)(12,0.07777778)
			};

            \tikzfillbetween[of=nonLinMin and nonLinMax]{blue, opacity=0.3};
            \tikzfillbetween[of=dledMin and dledMax]{violet, opacity=0.3};
            \tikzfillbetween[of=2dledMin and 2dledMax]{violet, opacity=0.3};

			\end{axis}
			\end{tikzpicture}
		}
		\caption{Original function}
		\label{sec:expInit:noiseAndVIGfill:raw}
	\end{subfigure}
    \begin{subfigure}[b]{0.49\linewidth}
		\resizebox{\linewidth}{!}{%
			\tikzset{every mark/.append style={scale=2.5}}
			\begin{tikzpicture}
			\begin{axis}[%
            legend entries={non-linear, non-monotnical, 2DLED},
            legend columns=-1,
            legend to name=named,
			%legend columns=-1,
            %legend entries={dgGA,cGOMEA,P3,P3-DLED,LT-GOMEA-DLED},
            %legend to name=named,
            %legend columns=-1,
            %legend entries={LT-GOMEA-DLED,LT-GOMEA-SLL,P3-DLED,P3-SLL},
            %legend to name=named,
			xtick={1,2,3,4,5,6,7,8,9,10,11,12},
			xticklabels={1.0, ,1.2, ,1.4, ,1.6, , 1.8, , 2.0, 2.5},
			xmin=1,
			xmax=12,
			%ymode=log,
			ymin=0,
			ymax=1,
			%legend pos=south east,
			xlabel=\textbf{maximum noise volume ($nVol$)},
			ylabel=\textbf{VIG epistasis (VIG fill)},
			grid,
			grid style=dashed,
			ticklabel style={scale=1.5},
			label style={scale=1.5},
			legend style={font=\fontsize{8}{0}\selectfont}
			]

            %\addplot[
			%color=brown,
			%mark=triangle*,
			%]
			%coordinates {
			%	(1,477026)(2,1961269)(3,7857276.5)(4,11729329)(5,15856165)(6,30264820)(7,24703945)(8,32664404)
			%};
   
            %\addplot[
			%color=black,
			%mark=*,
			%]
			%coordinates {
            %    (1,439884.5)(2,1464475.5)(3,4787594)(4,7602447)(5,13019899.5)(6,22470466)(7,27529834)(8,37681398)
			%};

            \addplot[
			color=blue,
            name path=nonLinMax,
            forget plot
			]
			coordinates {
				(1,0)(2,0)(3,0)(4,1)(5,1)(6,1)(7,1)(8,1)(9,1)(10,1)(11,1)(12,1)
			};
   
            \addplot[
			color=blue,
			mark=*,
            name path=nonLinMed
			]
			coordinates {
				(1,0)(2,0)(3,0)(4,0)(5,0)(6,0)(7,0)(8,0)(9,0)(10,0)(11,0)(12,1)
			};

            \addplot[
			color=blue,
            name path=nonLinMin,
            forget plot
			]
			coordinates {
				(1,0)(2,0)(3,0)(4,0)(5,0)(6,0)(7,0)(8,0)(9,0)(10,0)(11,0)(12,0)
			};

            \addplot[
			color=violet,
            name path=dledMax,
            forget plot
			]
			coordinates {
				(1,0)(2,0)(3,0)(4,0.11111111)(5,0.44444444)(6,0.22222222)(7,0.82222222)(8,0.71111111)(9,0.73333333)(10,0.55555556)(11,0.77777778)(12,0.68888889)

			};

            \addplot[
			color=violet,
			mark=triangle*,
            name path=dledMed
			]
			coordinates {
				(1,0)(2,0)(3,0)(4,0)(5,0)(6,0)(7,0)(8,0)(9,0)(10,0)(11,0)(12,0.2)

			};

            \addplot[
			color=violet,
            name path=dledMin,
            forget plot
			]
			coordinates {
				(1,0)(2,0)(3,0)(4,0)(5,0)(6,0)(7,0)(8,0)(9,0)(10,0)(11,0)(12,0)
			};

            \addplot[
			color=red,
            name path=2dledMax,
            forget plot
			]
			coordinates {
				(1,0)(2,0)(3,0)(4,0.05555556)(5,0.23333333)(6,0.11111111)(7,0.52222222)(8,0.44444444)(9,0.47777778)(10,0.31111111)(11,0.48888889)(12,0.46666667)

			};

            \addplot[
			color=red,
			mark=square*,
            name path=2dledMed
			]
			coordinates {
				(1,0)(2,0)(3,0)(4,0)(5,0)(6,0)(7,0)(8,0)(9,0)(10,0)(11,0)(12,0.116666665)

			};

            \addplot[
			color=red,
            name path=2dledMin,
            forget plot
			]
			coordinates {
				(1,0)(2,0)(3,0)(4,0)(5,0)(6,0)(7,0)(8,0)(9,0)(10,0)(11,0)(12,0)
			};

            \tikzfillbetween[of=nonLinMin and nonLinMax]{blue, opacity=0.3};
            \tikzfillbetween[of=dledMin and dledMax]{violet, opacity=0.3};
            \tikzfillbetween[of=2dledMin and 2dledMax]{violet, opacity=0.3};

			\end{axis}
			\end{tikzpicture}
		}
		\caption{Denoised surrogate}
		\label{sec:expInit:noiseAndVIGfill:denoised}
	\end{subfigure}

	\hspace{10.11 cm}
	\ref{named}
	%\ref{named2}

	\caption{The influence of noise volume on the VIG epistasis for various dependency checks (onemax problem). Lines with markers refer to the median. The bottom and upper lines of the same color refer to maximal and minimal values.}
	\label{fig:expInit:noiseAndVIGfill}
\end{figure}
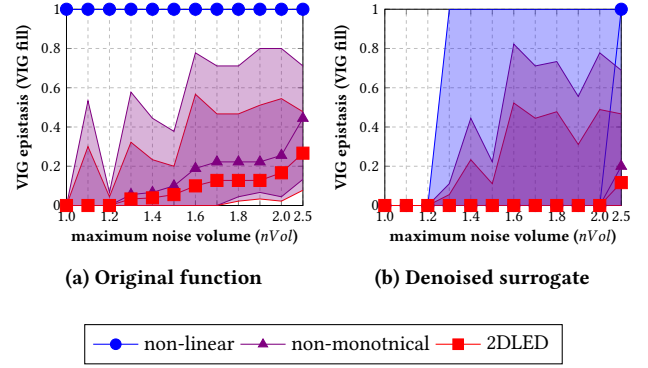

Figure \ref{sec:expInit:noiseAndVIGfill:raw} presents \textit{VIG epistasis} (the percentage of non-zero VIG entries) for the three considered dependency types for the noised onemax problem. In onemax, if two solutions refer to different function values, then the value of this difference is at least one. Thus, for the lowest considered noise, i.e., $nVol=1$, for any pair of solutions $\vec{x_a}$ and $\vec{x_b}$ such that $f(\vec{x_a}) < f(\vec{x_b})$ it is true that $f'(\vec{x_a}) < f'(\vec{x_b})$, where $f'(\vec{x})$ is the original function after adding noise. As presented in Figure \ref{sec:expInit:noiseAndVIGfill:raw}, the non-linear VIG becomes a full graph in all tested cases, even for the lowest noise values. Since onemax does not yield any dependencies, all discovered dependencies arise from noise and may be considered false. Using such dependencies may deteriorate the effectiveness of the optimizers using it \cite{linkageQuality}.\par

For DLED and 2DLED VIGs, even their maximal values for the highest noise ($nVal=2.5$) are not full graphs, and the median epistasis is below $50\%$ and $30$\%, respectively. This indicates that DLED and 2DLED may be useful in decomposing problem instances in which the optimized function has noise-like characteristics. \par

\begin{algorithm}
	\caption{Walsh denoise procedure}
	\begin{algorithmic}[1]
        \Function{LandVis}{$WalshCoeffsAll$}
            \State $\textit{CoeffsSorted} \gets$ SortByAbsValueAsc($\textit{WalshCoeffsAll}$); \label{line:expInit:denoise:sort}
            \State $\textit{CoeffsSurr} \gets \textit{CoeffsSorted}$;
            \State $\textit{OrigOptima} \gets $ GetAllOptima($\textit{CoeffsSorted}$);
            \State $\textit{SurrOptima} \gets $ GetAllOptima($\textit{CoeffsSurr}$);
            \While{$\textit{OrigOptima} = \textit{SurrOptima}$} \label{line:expInit:denoise:compare}
                \State $\textit{CoeffsSurrTry} \gets \textit{CoeffsSurr} - \textit{CoeffsSurr}[0]$; \label{line:expInit:denoise:remove}
                \State $\textit{SurrOptima} \gets $ GetAllOptima($\textit{CoeffsSurrTry}$);
                \If{$\textit{OrigOptima} = \textit{SurrOptima}$} \label{line:expInit:denoise:compareIf}
                    \State $\textit{CoeffsSurr} \gets \textit{CoeffsSurrTry}$;
                \EndIf
            \EndWhile
            \Return $CoeffsSurr$;
		\EndFunction
        \end{algorithmic}
	\label{alg:expInit:denoise}
\end{algorithm}

To show that the Walsh function representation may be useful for removing the noise, we propose the \textit{Walsh-based denoise procedure} (Pseudocode \ref{alg:expInit:denoise}) dedicated to analyzing toy-sized instances to understand their features. The use of knowledge gathered in this way will be the subject of future work. Considering their absolute value, we sort all Walsh coefficients in increasing order (line \ref{line:expInit:denoise:sort}). We remove subsequent coefficients from the function representation (line \ref{line:expInit:denoise:remove}) as long as, after removing a given coefficient, the global optima of the surrogate are the same as the global optima of the original function (lines \ref{line:expInit:denoise:compare} and \ref{line:expInit:denoise:compareIf}). Function representation obtained in this way will be denoted as \textit{denoised surrogate} or \textit{denoised function}.\par

As presented in Figure \ref{sec:expInit:noiseAndVIGfill:denoised}, the VIG epistasis of the denoised surrogate is significantly lower for all considered dependency types. For $nVol \leq 1.2$, median epistasis is zero, although in some runs, it may raise to $100$\% for $nVol \geq 1.3$ and non-linear VIG. Even for the highest considered noise, the epistasis of the denoised surrogate remains low for DLED and 2DLED VIGs.\par

\begin{table}[ht]
\caption{The influence of the mask size and noise volume on the median minimum absolute Walsh coefficient value.}
	\label{tab:expInit:minCoeffVal}
\scriptsize
\begin{tabular}{c|cccc}
\textbf{noise}  & \multicolumn{4}{c}{\textbf{Walsh coefficient mask size}}    \\
\textbf{volume ($nVol$)} & \textbf{2}    & \textbf{3}    & \textbf{4}    & \textbf{5}    \\
\hline
\textbf{1.0}    & \textbf{2.0E-4} & 4.2E-5 & 1.1E-7 & 1.5E-7 \\
\textbf{1.1}    & \textbf{1.5E-4} & 4.5E-7 & 7.0E-8 & 7.0E-8 \\
\textbf{1.2}    & \textbf{2.1E-4} & 4.3E-5 & 8.5E-8 & 3.0E-8 \\
\textbf{1.3}    & \textbf{3.6E-4} & 2.4E-5 & 8.0E-8 & 8.0E-8 \\
\textbf{1.4}    & \textbf{3.3E-4} & 8.6E-7 & 1.3E-7 & 1.3E-7 \\
\textbf{1.5}    & \textbf{3.1E-4} & 2.7E-5 & 1.1E-7 & 6.5E-8 \\
\textbf{1.6}    & \textbf{2.9E-4} & 6.5E-7 & 2.2E-7 & 7.5E-8 \\
\textbf{1.7}    & \textbf{3.4E-4} & 3.9E-5 & 1.5E-7 & 1.0E-7 \\
\textbf{1.8}    & \textbf{2.7E-4} & 7.9E-5 & 1.9E-7 & 5.0E-8 \\
\textbf{1.9}    & \textbf{3.7E-4} & 5.8E-5 & 1.7E-7 & 6.0E-8 \\
\textbf{2.0}    & \textbf{5.7E-4} & 1.1E-4 & 1.2E-7 & 3.6E-7 \\
\textbf{2.5}    & \textbf{4.4E-4} & 8.5E-5 & 1.5E-7 & 2.2E-7
\end{tabular}
\end{table}

%\begin{table}[]
%\caption{The influence of the mask size and noise volume on the median minimum absolute Walsh coefficient value.}
%	\label{tab:expInit:minCoeffVal}
%\scriptsize
%\begin{tabular}{c|cccccccc}
%\textbf{coeff} & \multicolumn{8}{c}{\textbf{noise volume ($nVol$)}} \\
%\textbf{mask size} & \textbf{1.0}   & \textbf{1.1}    & \textbf{1.2}    & \textbf{1.4}    & \textbf{1.6}    & \textbf{1.8}    & \textbf{2.0}    & \textbf{2.5}    \\
%\hline
%\textbf{2}   & 2.0E\textbf{-4} & 1.5E\textbf{-4} & 2.1E\textbf{-4} & 3.3E\textbf{-4} & 2.9E\textbf{-4} & 2.7E\textbf{-4} & 5.7E\textbf{-4} & 4.4E\textbf{-4} \\
%\textbf{3}   & 4.2E-5 & 4.5E-7 & 4.3E-5 & 8.6E-7 & 6.5E-7 & 7.9E-5 & 1.1E\textbf{-4} & 8.5E-5 \\
%\textbf{4}   & 1.1E-7 & 7.0E-8 & 8.5E-8 & 1.3E-7 & 2.2E-7 & 1.9E-7 & 1.2E-7 & 1.5E-7 \\
%\textbf{5}   & 1.5E-7 & 7.0E-8 & 3.0E-8 & 1.3E-7 & 7.5E-8 & 5.0E-8 & 3.6E-7 & 2.2E-7
%\end{tabular}
%\end{table}

In onemax, all Walsh coefficients with masks of size 2 or higher are zero. Thus, analyzing such coefficients for the noised onemax will show how the noise may influence those absolute coefficient values that refer to variable dependencies. The median and maximum values were similar for all mask sizes. Therefore, in Table \ref{tab:expInit:minCoeffVal}, we report the median of the minimal absolute Walsh coefficient values for masks of a given size. This value is significantly higher for Walsh coefficients referring to masks of size two. Thus, the denoise procedure will consider many Walsh coefficients with longer masks first. This indicates that the noise modelled by Walsh coefficients with masks of size two may be relatively hard to remove. Such observations are coherent with the research presented in \cite{walshSurBin}. \par

%Moreover, the dependency reduction techniques proposed \cite{superMasksGray} will not reduce the dependencies arising from such coefficients.\par

The results reported in this section indicate that the dependency checks based on non-monotonicity are more useful than the non-linearity check for decomposing problems with noise-like features. The proposed denoised procedure may be useful in limiting the number of dependencies caused by noise for sufficiently short problem instances. Finally, we show that Walsh coefficients referring to the masks of size 2, that model the noise may be the hardest to remove using the proposed denoised procedure. \par

\subsection{Denoising real-world instances}
\label{sec:expInit:denoise}

We consider two real-world feature selection problems, Bare Soil Detection in Remotely-sensed Earth Observation data~\cite{2021_Silvero, 2023_Majeed} and Anomaly Detection in Satellite Telemetry~\cite{bruszzz24sd_opssat}. Section S-I (supplementary material) highlights their importance and details. In both cases, compressing machine learning pipelines---by selecting discriminative features or optimizing models---is essential for practical deployment. Without such optimization, the models may become unusable, as they cannot be onboarded to the target environment.\par

We divide the problems tackled in our experimental study into three groups: kNN, BS (bare soil detection), and OPS (OPS-SAT telemetry analysis). Note that each of these groups considers many different quality measures. Thus, these are groups of different problems. Nevertheless, for presentation clarity, we will show the summarized results divided into these three groups.

\begin{table}[ht]
\caption{VIG epistasis of the considered real-world instances for various dependency checks after denoising procedure.}
	\label{tab:expInit:denoise:epistasis}
\scriptsize
\begin{tabular}{l|lll|lll|lll}
      & \multicolumn{3}{c}{\textbf{non-linearity}}  & \multicolumn{3}{c}{\textbf{non-monotonicity}} & \multicolumn{3}{c}{\textbf{2DLED}}       \\
      & \textbf{kNN} & \textbf{BS} & \textbf{OPS} & \textbf{kNN} & \textbf{BS} & \textbf{OPS}  & \textbf{kNN} & \textbf{BS} & \textbf{OPS}  \\
      \hline
\textbf{min}   & 0.00                & 1.00                & 0.27 & 0.00 & 0.42 & 0.05 & 0.00  & 0.38 & 0.03 \\
\textbf{max}   & 1.00                & 1.00                & 1.00 & 0.95 & 0.82 & 0.90 & 0.71  & 0.60 & 0.70 \\
\textbf{avr}   & 1.00                & 1.00                & 0.90 & 0.42 & 0.53 & 0.48 & 0.25  & 0.47 & 0.30 \\
\textbf{std} & 0.05                & 0.00                & 0.22 & 0.19 & 0.09 & 0.22 & 0.13  & 0.04 & 0.18 \\
\textbf{med}  & 1.00                & 1.00                & 1.00 & 0.37 & 0.51 & 0.46 & 0.22  & 0.47 & 0.25
\end{tabular}
\end{table}

For every considered real-world problem instance, all considered VIG types were full graphs. However, VIGs obtained after denoising differ significantly. Table \ref{tab:expInit:denoise:epistasis} shows that for the denoised surrogate, the non-linear VIG may be an empty graph (no dependent variable pairs). Nevertheless, most non-linear VIGs remain a full graph. Similarly to results presented for noised onemax, non-monotonicity check and 2DLED seem to be more useful, i.e., the VIG epistasis for these dependency types is significantly lower and does not reach $100$\% in any of the performed runs.\par

\begin{table}[ht]
\caption{The number of maximum full subgraphs for VIGs concenring various dependency checks for considered real-world instances.}
	\label{tab:expInit:denoise:glbs}
\scriptsize
\begin{tabular}{l|rr|rr|rr}
         & \multicolumn{2}{c}{\textbf{non-linearity}}  & \multicolumn{2}{c}{\textbf{non-monotonicity}}   & \multicolumn{2}{c}{\textbf{2DLED}}     \\
         & \textbf{Number*}            & \textbf{Len**}   & \textbf{Number}            & \textbf{Len}   & \textbf{Number}            & \textbf{Len}     \\
     %& \textbf{med} & \textbf{min/} & \textbf{med} & \textbf{min/} & \textbf{med} & \textbf{min/} \\
      %   & \textbf{med min/max} & \textbf{max} & \textbf{med min/max} & \textbf{max} & \textbf{med min/max} & \textbf{max} \\
         \hline
\textbf{kNN}    & 1 1/15  & 1/15  & 15 5/36  & 1/12  & 27 13/180 & 1/12  \\
\textbf{BS}  & 1 1/1   & 19/19  & 9 3/142 & 1/14  & 13 6/383 & 1/14  \\
\textbf{OPS}   & 1 1/23  & 1/18  & 25 9/46  & 1/13  & 58 21/239 & 1/13 \\
    \hline
  \multicolumn{7}{l}{* statistics in \textit{Number} column: med min/max}     \\
  \multicolumn{7}{l}{** statistics in \textit{Len} column: min/max}  
\end{tabular}
\end{table}

In Table \ref{tab:expInit:denoise:glbs}, we present the number of maximal full subgraphs (cliques) obtained for VIGs of various types constructed for denoised surrogates. Maximal full subgraphs of a VIG can be interpreted as sets of entry variables to subfunctions in the additive form \cite{maximalFullSubgraphs}. Thus, if the problem has a strong structure it should decompose into many subfunctions (overlapping or not). Table \ref{tab:expInit:denoise:glbs} shows that for the non-linear VIG, the median number of maximal full subgraphs is~$1$ (because the median epistasis of non-linear VIG is $100$\%). However, even for non-linear VIG, in some cases, the number of maximal full subgraphs is equal to the number of function arguments, which shows that after denoising, we deal with a fully separable problem in which each variable can be optimized separately.\par

%In Table \ref{tab:expInit:denoise:glbs}, we present the number of maximal full subgraphs (cliques) obtained for VIGs of various types constructed for denoised surrogates. Maximal full subgraphs of a VIG can be interpreted as sets of entry variables to subfunctions in the additive form \cite{maximalFullSubgraphs}. Thus, if the problem has a strong structure it should decompose into many subfunctions (overlapping or not). As presented in Table \ref{tab:expInit:denoise:glbs}, for the non-linear VIG, the median number of maximal full subraphs is $1$ (because the median epistasis of non-linear VIG is $100$\%). However, even for non-linear VIG, in some cases, the number of maximal full subgraphs is equal to the number of function arguments, which shows that after denoising, we deal with a fully separable problem, in which each variable can be optimized separately.\par

%For DLED and 2DLED VIGs, the number of maximal full subgraphs is high. It sometimes exceeds the number of function variables, which shows that the denoised problem is built from overlapping subfunctions. Finally, for DLED and 2DLED VIGs, the lowest number of maximal full subgraphs is always higher than one. Thus, we may state that all of the considered instances have a decomposable structure hidden under the noise.\par

For non-monotonicity and 2DLED VIGs, the number of maximal full subgraphs is high. It sometimes exceeds the number of function variables, which shows that the denoised problem is built from overlapping subfunctions. Finally, for these VIGs, the lowest number of maximal full subgraphs is always higher than one. Thus, we can state that all of the considered instances have a decomposable structure hidden under the noise.\par

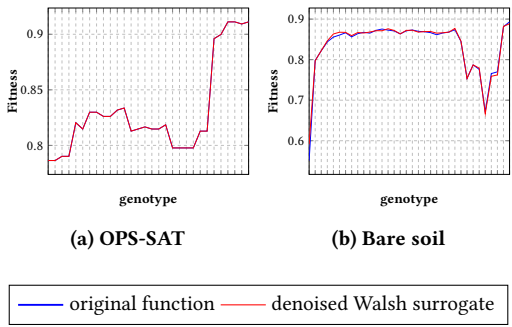
\begin{figure}[ht]
    \begin{subfigure}[b]{0.40\linewidth}
		\resizebox{\linewidth}{!}{%
			\tikzset{every mark/.append style={scale=2.5}}
			\begin{tikzpicture}
			\begin{axis}[%
            legend entries={original function, denoised Walsh surrogate},
            legend columns=-1,
            legend to name=named,
			%legend columns=-1,
            %legend entries={dgGA,cGOMEA,P3,P3-DLED,LT-GOMEA-DLED},
            %legend to name=named,
            %legend columns=-1,
            %legend entries={LT-GOMEA-DLED,LT-GOMEA-SLL,P3-DLED,P3-SLL},
            %legend to name=named,
			xtick={1,2,3,4,5,6,7,8,9,10,11,12,13,14,15,16,17,18,19,20,21,22,23,24,25,26,27,28,29,30},
			xticklabels={},
			xmin=1,
			xmax=30,
			%ymode=log,
			%ymin=0,
			%ymax=1,
			%legend pos=south east,
			xlabel=\textbf{genotype},
			ylabel=\textbf{Fitness},
			grid,
			grid style=dashed,
			ticklabel style={scale=1.5},
			label style={scale=1.5},
			legend style={font=\fontsize{8}{0}\selectfont}
			]

            %\addplot[
			%color=brown,
			%mark=triangle*,
			%]
			%coordinates {
			%	(1,477026)(2,1961269)(3,7857276.5)(4,11729329)(5,15856165)(6,30264820)(7,24703945)(8,32664404)
			%};
   
            %\addplot[
			%color=black,
			%mark=*,
			%]
			%coordinates {
            %    (1,439884.5)(2,1464475.5)(3,4787594)(4,7602447)(5,13019899.5)(6,22470466)(7,27529834)(8,37681398)
			%};

            \addplot[
			color=blue,
			]
			coordinates {
                (1,0.78639)(2,0.78639)(3,0.79017)(4,0.79017)(5,0.82042)(6,0.81474)(7,0.82987)(8,0.82987)(9,0.82609)(10,0.82609)(11,0.83176)(12,0.83365)(13,0.81285)(14,0.81474)(15,0.81664)(16,0.81474)(17,0.81474)(18,0.81853)(19,0.79773)(20,0.79773)(21,0.79773)(22,0.79773)(23,0.81285)(24,0.81285)(24,0.81285)(25,0.89603)(26,0.89981)(27,0.91115)(28,0.91115)(29,0.90926)(30,0.91115)

			};

            \addplot[
			color=red
			]
			coordinates {
				(1,0.78639002)(2,0.78638999)(3,0.79017001)(4,0.79016999)(5,0.82042)(6,0.81473998)(7,0.82987001)(8,0.82987001)(9,0.82608999)(10,0.82609)(11,0.83176002)(12,0.83365001)(13,0.81285002)(14,0.81474)(15,0.81663998)(16,0.81474003)(17,0.81474003)(18,0.81853)(19,0.79772999)(20,0.79772998)(21,0.79773001)(22,0.79773)(23,0.81284997)(24,0.81284999)(24,0.81284999)(25,0.89603001)(26,0.89980999)(27,0.91114998)(28,0.91114998)(29,0.90926003)(30,0.91115002)

			};

			\end{axis}
			\end{tikzpicture}
		}
		\caption{OPS-SAT}
		\label{fig:expInit:denoise:landscape:opsSat}
	\end{subfigure}
    \begin{subfigure}[b]{0.40\linewidth}
		\resizebox{\linewidth}{!}{%
			\tikzset{every mark/.append style={scale=2.5}}
			\begin{tikzpicture}
			\begin{axis}[%
            legend entries={original function, denoised Walsh surrogate},
            legend columns=-1,
            legend to name=named,
			%legend columns=-1,
            %legend entries={dgGA,cGOMEA,P3,P3-DLED,LT-GOMEA-DLED},
            %legend to name=named,
            %legend columns=-1,
            %legend entries={LT-GOMEA-DLED,LT-GOMEA-SLL,P3-DLED,P3-SLL},
            %legend to name=named,
			xtick={1,2,3,4,5,6,7,8,9,10,11,12,13,14,15,16,17,18,19,20,21,22,23,24,25,26,27,28,29,30,31,32,33,34},
			xticklabels={},
			xmin=1,
			xmax=34,
			%ymode=log,
			%ymin=0,
			%ymax=1,
			%legend pos=south east,
			xlabel=\textbf{genotype},
			ylabel=\textbf{Fitness},
			grid,
			grid style=dashed,
			ticklabel style={scale=1.5},
			label style={scale=1.5},
			legend style={font=\fontsize{8}{0}\selectfont}
			]

            %\addplot[
			%color=brown,
			%mark=triangle*,
			%]
			%coordinates {
			%	(1,477026)(2,1961269)(3,7857276.5)(4,11729329)(5,15856165)(6,30264820)(7,24703945)(8,32664404)
			%};
   
            %\addplot[
			%color=black,
			%mark=*,
			%]
			%coordinates {
            %    (1,439884.5)(2,1464475.5)(3,4787594)(4,7602447)(5,13019899.5)(6,22470466)(7,27529834)(8,37681398)
			%};

            \addplot[
			color=blue,
            line width=0.25mm
			]
			coordinates {
				(1,0.55115588)(2,0.7981781)(3,0.82187562)(4,0.84394456)(5,0.8556548)(6,0.86060518)(7,0.86630448)(8,0.85623836)(9,0.86399128)(10,0.86740226)(11,0.86534464)(12,0.87214293)(13,0.875489)(14,0.8728081)(15,0.87077271)(16,0.86346497)(17,0.87109653)(18,0.87310649)(19,0.86981985)(20,0.86878)(21,0.86656817)(22,0.86140774)(23,0.8657157)(24,0.86781328)(24,0.86781328)(25,0.87382098)(26,0.84501657)(27,0.75376429)(28,0.78696993)(29,0.7761766)(30,0.67140547)(31,0.76562943)(32,0.76991707)(33,0.88128831)(34,0.89315646)
			};

            \addplot[
			color=red
			]
			coordinates {
				(1,0.59142388)(2,0.79722562)(3,0.82217075)(4,0.84658572)(5,0.8632989)(6,0.86762549)(7,0.86722059)(8,0.85925168)(9,0.86679492)(10,0.86564738)(11,0.86805826)(12,0.87138492)(13,0.87122636)(14,0.87618751)(15,0.8719825)(16,0.86348009)(17,0.87213873)(18,0.87257621)(19,0.867798)(20,0.86922899)(21,0.86939153)(22,0.86530896)(23,0.86630603)(24,0.86791483)(24,0.86791483)(25,0.87697929)(26,0.84558575)(27,0.75348552)(28,0.787074)(29,0.77907958)(30,0.66676996)(31,0.75889303)(32,0.76231206)(33,0.88184772)(34,0.88858144)
			};

			\end{axis}
			\end{tikzpicture}
		}
		\caption{Bare soil}
		\label{fig:expInit:denoise:landscape:bareSoil}
	\end{subfigure}

	\hspace{10.11 cm}
	\ref{named}
	%\ref{named2}

	\caption{A landscape cross-section for chosen problem instances (solution opposite to global optimum $\rightarrow$ global minimum $\rightarrow$ global maximum; differing genes are modified in a random order). See Fig. S-1, supplementary material, for more results.}
	\label{fig:expInit:denoise:landscape}
\end{figure}

\begin{figure*}[ht]
    \begin{subfigure}[b]{0.19\linewidth}
		\resizebox{\linewidth}{!}{%
			\tikzset{every mark/.append style={scale=2.5}}
			\begin{tikzpicture}
			\begin{axis}[%
            legend entries={LT-GOMEA, P3, GBO-PHE-PX, GBO-PHE-LTop-wdVIG},
            legend columns=-1,
            legend to name=named,
			%legend columns=-1,
            %legend entries={dgGA,cGOMEA,P3,P3-DLED,LT-GOMEA-DLED},
            %legend to name=named,
            %legend columns=-1,
            %legend entries={LT-GOMEA-DLED,LT-GOMEA-SLL,P3-DLED,P3-SLL},
            %legend to name=named,
			xtick={1,2,3,4,5},
			xticklabels={1,2,3,4,5},
			xmin=0.5,
			xmax=5.5,
			%ymode=log,
			%ymin=1e4,
			%ymax=1e9,
			%legend pos=south east,
			%xlabel=\textbf{# of random Walsh coeff per var.},
			ylabel=\textbf{FFE until opt.},
			grid,
			grid style=dashed,
			ticklabel style={scale=1.5},
			label style={scale=1.5},
			legend style={font=\fontsize{8}{0}\selectfont}
			]

                \addplot[
			color=black,
			mark=square*,
			]
			coordinates {
				(1,407994)
			};

            \addplot[
			color=blue,
			mark=triangle*,
			]
			coordinates {
				(1,129564)(2,281529)(3,395659)
			};

			\addplot[
			color=darkgreen,
			mark=square*,
			]
			coordinates {
                (1,590335)
			};

                \addplot[
			color=red,
			mark=*,
			]
                coordinates {
                (1,205610.5)(2,190572)(3,154626)(4,224153.5)(5,120081)
			};
			
			\end{axis}
			\end{tikzpicture}
		}
		\caption{\textit{NKLand6k ($n=72$)}}
		\label{fig:resScala:NKLand6k}
	\end{subfigure}
    \begin{subfigure}[b]{0.19\linewidth}
		\resizebox{\linewidth}{!}{%
			\tikzset{every mark/.append style={scale=2.5}}
			\begin{tikzpicture}
			\begin{axis}[%
            legend entries={LT-GOMEA, P3, GBO-PHE-PX, GBO-PHE-LTop-wdVIG},
            legend columns=-1,
            legend to name=named,
			%legend columns=-1,
            %legend entries={dgGA,cGOMEA,P3,P3-DLED,LT-GOMEA-DLED},
            %legend to name=named,
            %legend columns=-1,
            %legend entries={LT-GOMEA-DLED,LT-GOMEA-SLL,P3-DLED,P3-SLL},
            %legend to name=named,
			xtick={1,2,3,4,5},
			xticklabels={1,2,3,4,5},
			xmin=0.5,
			xmax=5.5,
			%ymode=log,
			ymin=1e5,
			ymax=8e5,
			%legend pos=south east,
			%xlabel=\textbf{unitation},
			ylabel=\textbf{FFE until opt.},
			grid,
			grid style=dashed,
			ticklabel style={scale=1.5},
			label style={scale=1.5},
			legend style={font=\fontsize{8}{0}\selectfont}
			]

                \addplot[
			color=red,
			mark=*,
			]
                coordinates {
                (1,275361)(2,352431)(3,238581.5)(4,311044)(5,322713)

			};
			
			\end{axis}
			\end{tikzpicture}
		}
		\caption{\textit{ISG ($n=256$)}}
		\label{fig:resScala:ISG}
	\end{subfigure}
    \begin{subfigure}[b]{0.19\linewidth}
		\resizebox{\linewidth}{!}{%
			\tikzset{every mark/.append style={scale=2.5}}
			\begin{tikzpicture}
			\begin{axis}[%
            legend entries={LT-GOMEA, P3, GBO-PHE-PX, GBO-PHE-LTop-wdVIG},
            legend columns=-1,
            legend to name=named,
			%legend columns=-1,
            %legend entries={dgGA,cGOMEA,P3,P3-DLED,LT-GOMEA-DLED},
            %legend to name=named,
            %legend columns=-1,
            %legend entries={LT-GOMEA-DLED,LT-GOMEA-SLL,P3-DLED,P3-SLL},
            %legend to name=named,
			xtick={1,2,3,4,5},
			xticklabels={1,2,3,4,5},
			xmin=0.5,
			xmax=5.5,
			%ymode=log,
			%ymin=1e4,
			%ymax=1e9,
			%legend pos=south east,
			%xlabel=\textbf{unitation},
			ylabel=\textbf{FFE until opt.},
			grid,
			grid style=dashed,
			ticklabel style={scale=1.5},
			label style={scale=1.5},
			legend style={font=\fontsize{8}{0}\selectfont}
			]
			
			\addplot[
			color=black,
			mark=square*,
			]
			coordinates {
				(1,656995.5)(2,854650)
			};

            \addplot[
			color=blue,
			mark=triangle*,
			]
			coordinates {
				(1,283400)(2,351478.5)(3,402515.5)(4,703840.5)(5,752771)
			};

                \addplot[
			color=red,
			mark=*,
			]
                coordinates {
                (1,218443.5)(2,220271.5)(3,184640.5)(4,218184.5)(5,218081.5)
			};
			
			\end{axis}
			\end{tikzpicture}
		}
		\caption{\textit{Dec8 ($n=104$)}}
		\label{fig:resScala:dec8}
	\end{subfigure}
    \begin{subfigure}[b]{0.19\linewidth}
		\resizebox{\linewidth}{!}{%
			\tikzset{every mark/.append style={scale=2.5}}
			\begin{tikzpicture}
			\begin{axis}[%
            legend entries={LT-GOMEA, P3, GBO-PHE-PX, GBO-PHE-LTop-wdVIG},
            legend columns=-1,
            legend to name=named,
			%legend columns=-1,
            %legend entries={dgGA,cGOMEA,P3,P3-DLED,LT-GOMEA-DLED},
            %legend to name=named,
            %legend columns=-1,
            %legend entries={LT-GOMEA-DLED,LT-GOMEA-SLL,P3-DLED,P3-SLL},
            %legend to name=named,
			xtick={1,2,3,4,5},
			xticklabels={1,2,3,4,5},
			xmin=0.5,
			xmax=5.5,
			%ymode=log,
			ymin=1e5,
			ymax=4e5,
			%legend pos=south east,
			%xlabel=\textbf{unitation},
			ylabel=\textbf{FFE until opt.},
			grid,
			grid style=dashed,
			ticklabel style={scale=1.5},
			label style={scale=1.5},
			legend style={font=\fontsize{8}{0}\selectfont}
			]

                \addplot[
			color=red,
			mark=*,
			]
                coordinates {
                (1,142222.5)(2,145208.5)(3,109772)(4,150182.5)(5,129367.5)
			};
			
			\end{axis}
			\end{tikzpicture}
		}
		\caption{\textit{dec8o5 ($n=45$)}}
		\label{fig:resScala:dec8o5}
	\end{subfigure}
    \begin{subfigure}[b]{0.19\linewidth}
		\resizebox{\linewidth}{!}{%
			\tikzset{every mark/.append style={scale=2.5}}
			\begin{tikzpicture}
			\begin{axis}[%
            legend entries={LT-GOMEA, P3, GBO-PHE-PX, GBO-PHE-LTop-wdVIG},
            legend columns=-1,
            legend to name=named,
			%legend columns=-1,
            %legend entries={dgGA,cGOMEA,P3,P3-DLED,LT-GOMEA-DLED},
            %legend to name=named,
            %legend columns=-1,
            %legend entries={LT-GOMEA-DLED,LT-GOMEA-SLL,P3-DLED,P3-SLL},
            %legend to name=named,
			xtick={1,2,3,4,5},
			xticklabels={1,2,3,4,5},
			xmin=0.5,
			xmax=5.5,
			%ymode=log,
			%ymin=1e4,
			%ymax=1e9,
			%legend pos=south east,
			%xlabel=\textbf{unitation},
			ylabel=\textbf{FFE until opt.},
			grid,
			grid style=dashed,
			ticklabel style={scale=1.5},
			label style={scale=1.5},
			legend style={font=\fontsize{8}{0}\selectfont}
			]
			
			\addplot[
			color=black,
			mark=square*,
			]
			coordinates {
				(1,167487)(2,201358.5)(3,230709.5)(4,390200.5)(5,478909)
			};

            \addplot[
			color=blue,
			mark=triangle*,
			]
			coordinates {
				(1,87836)(2,101341)(3,82913.5)(4,111449.5)(5,146622)
			};

			\addplot[
			color=darkgreen,
			mark=square*,
			]
			coordinates {
                (1,594993)
			};

                \addplot[
			color=red,
			mark=*,
			]
                coordinates {
                (1,33316.5)(2,31246)(3,40304)(4,32325.5)(5,32514)
			};
			
			\end{axis}
			\end{tikzpicture}
		}
		\caption{\textit{bim6o4 ($n=100$)}}
		\label{fig:resScala:bim6o4}
	\end{subfigure}

	\hspace{10.11 cm}
	\ref{named}
	%\ref{named2}

	\caption{Scalablity for chosen instances and increasing noise (X axis: number of random Walsh coefficients per variable).}
	\label{fig:resScala}
\end{figure*}
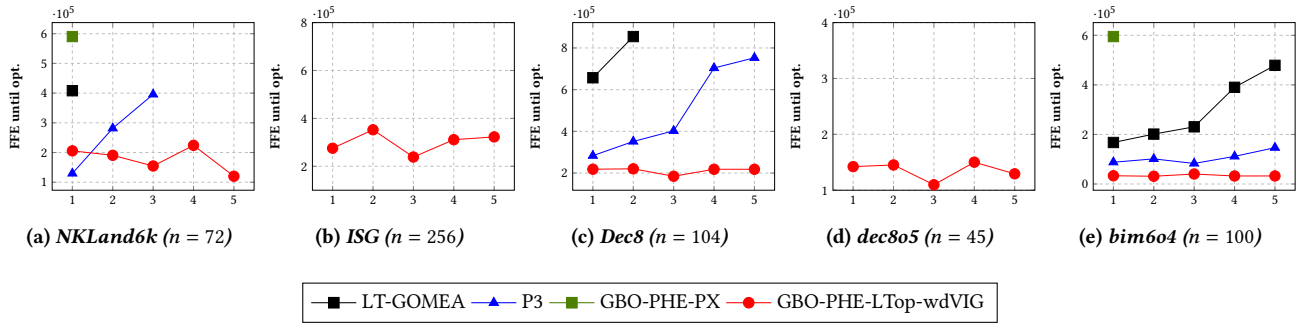[ht]

\begin{table*}[ht]
\caption{Maximum problem size for which a given optimizer found an optimal solution in at least 80\% of the runs. GBO-PHE-LTop-wdVIG was the most effective. Therefore, we compare all other optimizers with this version of GBO-PHE.}
	\label{tab:res:gen}
\scriptsize
\begin{tabular}{lc|rr|rrrrrrrrr}
       &       &     &          &  \multicolumn{9}{c}{\textbf{GBO-PHE}}          \\
       &       &  &  & \textbf{stand} & \textbf{Top}          & \textbf{Bottom}       & \textbf{Top}   & \textbf{Bottom} & \textbf{Top}     & \textbf{Bottom}  & \textbf{Top}     & \textbf{Bottom}  \\
       & \textbf{noise} & \textbf{P3}    &   \textbf{LT-GOMEA}       &  \textbf{PX}          & \textbf{wdVIG}        & \textbf{wdVIG}        & \textbf{wsVIG} & \textbf{wsVIG}  & \textbf{wdVIGns} & \textbf{wdVIGns} & \textbf{wsVIGns} & \textbf{wsVIGns} \\
       \hline
\textbf{nkLand6k}  & 0     & 102 & 72       & \textbf{204}        & \textbf{204}          & 150          & 48    & 24     & 150     & 150     & 24      & 48      \\
\textbf{nkLand6k}  & 5     & 48  & 48       & 24         & \textbf{150}          & \textbf{150}          & 24    & 48     & \textbf{150}     & 102     & 24      & 48      \\
\textbf{ISG}    & 0     & N/A & N/A      & 625        & \textbf{784}          & 256          & N/A   & N/A    & \textbf{784}     & 256     & N/A     & N/A     \\
\textbf{ISG}    & 5     & N/A & N/A      & N/A        & \textbf{400}          & 256          & N/A   & N/A    & \textbf{400}     & 256     & N/A     & N/A     \\
\textbf{bim6}   & 0     & \textbf{204} & \textbf{204}      & \textbf{204}        & \textbf{204}          & \textbf{204}          & \textbf{204}   & \textbf{204}    & \textbf{204}     & \textbf{204}     & \textbf{204}     & \textbf{204}     \\
\textbf{bim6}   & 5     & \textbf{204} & \textbf{204}      & 30         & \textbf{204}          & \textbf{204}          & 12    & \textbf{204}    & \textbf{204}     & \textbf{204}     & 12      & \textbf{204}     \\
\textbf{bim10}   &  0  & 60 & \textbf{100} & 60  & 50 & 50 & 30 & 30  & 30 & 20 & 20 & 30  \\
\textbf{bim10} & 5 & 30 & \textbf{50}  & N/A & \textbf{50} & 20 & 20 & N/A & 30 & 20 & 20 & N/A \\
\textbf{bim6o4} & 0     & \textbf{200} & \textbf{200}      & \textbf{200}        & \textbf{200}          & \textbf{200}          & 16    & \textbf{200}    & \textbf{200}     & \textbf{200}     & 16      & \textbf{200}     \\
\textbf{bim6o4} & 5     & \textbf{200} & 150      & 30         & \textbf{200}          & \textbf{200}          & 16    & \textbf{200}    & \textbf{200}     & \textbf{200}     & 16      & \textbf{200}     \\
\textbf{dec8} & 0     & \textbf{200} & 152      & \textbf{200}        & \textbf{200}          & 152          & \textbf{200}   & 152    & \textbf{200}     & 152     & \textbf{200}     & 104     \\
\textbf{dec8} & 5     & 104 & 16       & 16         & \textbf{152}          & \textbf{152}          & 16    & 104    & \textbf{152}     & \textbf{152}     & 16      & 104     \\
\textbf{dec8o5} & 0     & 15  & 15       & \textbf{60}         & \textbf{60}           & 15           & 15    & 15     & \textbf{60}      & 15      & 15      & 15      \\
\textbf{dec8o5} & 5     & 15  & 15       & 15         & \textbf{60}           & 15           & 15    & 15     & \textbf{60}      & 15      & 15      & 15  \\
\hline
\multicolumn{2}{r|}{\textbf{\textit{better than GBO-PHE-LTop-wdVIG}}} & 1 & 1 & 1  & \textbf{N/A}  &  0 & 0 & 0 & 0 & 0 & 0 & 0\\
\multicolumn{2}{r|}{\textbf{\textit{worse than GBO-PHE-LTop-wdVIG}}} & 8 & 9 & 8  & \textbf{N/A}  &  7 & 12 & 10 & 3 & 10 & 12 & 10
\end{tabular}
\end{table*}

%The proposed denoising procedure assumes that Walsh coefficients from the Walsh-based function model should be removed in ascending order, considering their absolute value. The denoised surrogate is acceptable as long as all globally optimal solutions of the original function are the same as all globally optimal solutions of the surrogate. Such a procedure seems convincing because it preserves the most relevant Walsh coefficients and most relevant features (the global optima set) of the optimized function.\par

To compare the landscapes of the original and denoised function we construct graphs in the following manner (for details, see Section S-III and Pseudocode S-1, supplementary material). We randomly choose one of the global optima and one of the global minima. We start from the solution opposite to the chosen global optimum, \textit{move} to the chosen global minimum, and \textit{move} to the chosen global maximum. By \textit{moving}, we consider flipping the genes differing in start and destination solutions. Figure \ref{fig:expInit:denoise:landscape} shows two representative curves. All curves for the original function have the same shape as the curves obtained for the denoised surrogates.\par

The results presented in this section lead to the following conclusions. Adding the noise to the optimized function may cause its VIG to become a full graph. VIGs based on the non-monotonicity check and 2DLED are significantly less vulnerable to discovering the dependencies caused by noise. The denoised surrogates of all the considered real-world instances are characterized by strong structure, even though for their original forms, we obtain full graphs for all considered VIG types. Thus, using weighted VIGs to identify and eliminate at least some part of the weakest dependencies is promising.

\section{Main Experiments}
\label{sec:expMain}

\subsection{Experiment setup and considered optimizers}
\label{sec:expMain:setup}

To verify the quality of the proposed wPX, we consider a set of benchmarks typical for the GA-related research, which includes the concatenations of: standard deceptive functions of order 8 \cite{decFunc} (denoted as \textit{dec8}), bimodal deceptive functions of order 6 and 10 \cite{decBimodalOld} (\textit{bim6} and \textit{bim10}). We also consider the cyclic traps built from these functions with various overlap size (e.g., dec8 with 5 overlapping genes for both neighbouring functions is denoted as \textit{dec8o5}). Finally, the set of problems was supplemented by NK-fitness landscapes of $k=6$ \cite{P3Original} (\textit{nkLand6k}), and Ising Spin Glasses \cite{P3Original,isg} (\textit{ISG}). The detailed definitions of these problems can be found in the supplementary material in Section S-IV.\par

All problems were considered in their standard and noised versions. The results presented in Section \ref{sec:expInit:noise} show that the noised-caused Walsh coefficients with masks of size 2 are the hardest to remove because their minimal absolute value is significantly higher than the minimal absolute value of coefficients with larger masks. Here, we consider problem instances that can be considered as large. Thus, adding random noise to each solution in the manner employed in Section \ref{sec:expInit:noise} is impossible. To simulate the noise, we randomly add the Walsh coefficient with a mask of size 2 for each variable, joining it with a randomly chosen variable. The absolute value of such a coefficient is small enough not to change the majority/minority relation between the fitness of any two solutions. However, two solutions with equal fitness in the original function can become unequal after adding noise. We consider adding 0, 1, 2, 3, 4, and 5 randomly generated Walsh coefficients of size 2 per a single variable (adding 0 coefficients refers to the original function).\par

In Section \ref{sec:PXwdVIG:GBOphe}, we propose GBO-PHE using wPX that employs wdVIG and the LTop strategy. To justify the intuitions behind these choices, we also consider GBO-PHE using \textit{LBot} strategy (denoted as \textit{GBO-PHE-LBot}) that considers all LT nodes in the P3 manner (the shorter nodes go first, the order of the nodes of the same size is random). Additionally, we consider other weighted VIG versions considered in this work, i.e., wdVIGns, wsVIG, and wsVIGns. Finally, the competing optimizer set is supplemented by GBO-PHE using standard PX (\textit{GBO-PHE-PX}). The considered optimizer set was supplemented by P3 and LT-GOMEA, which are black-box optimizers. To make the comparison fair, they were supported a VIG and did not use any problem decomposition mechanisms.\par

All the considered optimizers are parameter-less and no tuning was necessary. The computation budget was set to $2\cdot10^6$ fitness function evaluations (FFE). The gray-box settings refer only to access to Walsh-based function representation, no partial evaluations were used. Each experiment was repeated 30 times. The source code of all optimizers was joined in one C++ project and is available on Zenodo~\cite{zenodo-package} and GitHub\footnote{\url{https://github.com/przewooz/wVIG}} (with the results and setting files).

\subsection{Results}
\label{sec:expMain:results}

In Table \ref{tab:res:gen}, we present the general comparison of the considered optimizers. For each problem, we report the largest problem size for which a given optimizer has an optimal solution in at least 80\% of the runs. 
%More detailed results can be found in Section \textcolor{red}{S-XX} (supplementary material). 
If two optimizers are successful for the same problem's size, we consider such a result as a tie. We avoid further comparing which of them obtained a given result faster (in terms of FFE) because the differences seem negligible, and even if statistically significant, they would not show true supremacy.\par

As presented, GBO-PHE-LTop-wdVIG was the most effective optimizer. It was outperformed only for bim10. For all the other problems, it was able to optimize the largest problem instances, regardless of whether they were noised or not. The effectiveness of GBO-PHE-LBot-wdVIG was worse because this GBO-PHE version considers a higher number of masks, which, for some problems, was too expensive. The effectiveness of GBO-PHE-LTop-wdVIGns was the same as GBO-PHE-LBot-wdVIG for many problems. Such results show that dynamically adjusting the considered Walsh Coefficients and using the LTop strategy was more significant than considering the mask size influence. Nevertheless, this latter mechanism also positively influences the quality of the results.\par

The comparison with P3, LT-GOMEA, and GBO-PHE using standard PX was decisive. These optimizers can not effectively handle problems with noise because they are using too large masks. For some problems (e.g., dec8o5), the considered gray-box versions of P3 and LT-GOMEA performed worse than expected. This situation is because their black-box versions discover dependencies and create different LTs during the run. Here, they were using true but static VIG, which deteriorated their effectiveness in some cases. Such a result is justified because when solving overlapping problems, using many diverse LTs is beneficial for these optimizers \cite{3lo}. \par

The above observations are confirmed by the scalability analysis presented in Fig. \ref{fig:resScala}. Note that the influence of increasing noise is negligible for GBO-PHE-LTop-wdVIG. At the same time, the effectiveness of the traditional optimizers deteriorates significantly.\par

\section{Conclusions}
\label{sec:conc}
In this work, we propose and verify experimentally various weighted VIGs created on the base of Walsh coefficients. Proposed mechanisms allow the effective optimization of problems with and without noise. However, the proposed ideas and their experimental verifications may have a broader meaning. Since we model the noise by adding random-based Walsh coefficients to the problem definition, we can generate instances with the increasing influence of noise and verify which mechanisms remain effective in such scenarios. The proposed wdVIG extends the idea of using Walsh decomposition to construct VIG without deteriorating existing VIG advantages. For problems without noise, wdVIG is equivalent to VIG, but for noised problems, its advantage over traditional VIG is significant.\par

The proposed experiments show that for real-world problem instances that are currently considered to have a weak problem structure, there is a potential to propose a mechanism that can uncover their strong structure. Here, we only show the existence of a strong structure in the toy-sized problem instances. However, proposing denoising mechanisms for large-scale instances is the most important direction for future work. It can enable using gray-box operators such as PX for problems they currently do not apply.

\begin{acks}
We want to thank Darrell Whitley for the valuable discussions on Walsh decomposition, which were one of the inspirations for writing this work.\par

The work was supported by: Polish National Science Centre (NCN), Grant 2022/45/B/ST6/04150 (Michal Przewozniczek); University of Malaga, project PAR 4/2023 (Francisco Chicano); National Council for Scientific and Technological Development, CNPq grant \#306689/ 2021-9 and FAPESP grant \#2024/15430-5 (Renato Tin\'os); European Space Agency (ESA), projects 40001373339/22/NL/GLC/ov and 4000141301; FESL.10.25-IZ.01-07G5/23 (Agata Wijata); Silesian University of Technology (Jakub Nalepa); Opole University of Technology, DELTA project no. 272/24 (Bogdan Ruszczak).
\end{acks}

%The work was supported by the projects: 40001373339/22/NL/GLC/ov and 4000141301 funded by European Space Agency (ESA), and by GENESIS-2 (supported by $\Phi$-Lab, ESA). BR was supported by the the Opole University of Technology as part of the DELTA project no. 272/24. AMW was supported by the European Funds for Silesia 2021-2027 Program co-financed by the Just Transition Fund—project entitled ``Development of the Silesian biomedical engineering potential in the face of the challenges of the digital and green economy (BioMeDiG)'' (FESL.10.25-IZ.01-07G5/23).

% This work was supported by the projects: 40001373339/22/NL/GLC/ov and 4000141301 funded by the European Space Agency (ESA), and by GENESIS-2 (funded and supported by $\Phi$-Lab, ESA). BR was supported by the the Opole University of Technology as part of the DELTA project no. 272/24. JN and AMW were supported by the Silesian University of Technology grant for maintaining and developing research potential and the European Funds for Silesia 2021-2027 Program co-financed by the Just Transition Fund—project entitled ''Development of the Silesian biomedical engineering potential in the face of the challenges of the digital and green economy (BioMeDiG)'' (FESL.10.25-IZ.01-07G5/23).
	
	\bibliographystyle{ACM-Reference-Format}
	\bibliography{wVIG} 
	
\end{document}